\pdfoutput=1
\documentclass[11pt]{article}
\usepackage[]{acl}
\usepackage{times}
\usepackage{latexsym}
\usepackage[T1]{fontenc}
\usepackage[utf8]{inputenc}
\usepackage{microtype}
\usepackage{inconsolata}
\usepackage{graphicx}
\usepackage{caption}
\usepackage{subcaption}
\usepackage{amsmath,amsfonts,amssymb}
\usepackage{textcomp}
\usepackage{url}
\usepackage{booktabs}
\usepackage{multirow}
\usepackage{arydshln}

\newcommand{\ve}[1]{\boldsymbol{#1}}
\newcommand{\ma}[1]{\mathbf{#1}}
\newcommand{\seq}[1]{\{#1\}}
\newcommand{\set}[1]{\mathbb{#1}}
\title{Improving Language Transfer Capability of Decoder-only Architecture in Multilingual Neural Machine Translation}

\author{
\textbf{Zhi Qu$^\dag$ }
\textbf{Yiran Wang$^\ddagger$ }
\textbf{Chenchen Ding$^\dag$$^\ddagger$}
\\
\textbf{Hideki Tanaka$^\ddagger$ }
\textbf{Masao Utiyama$^\ddagger$ }
\textbf{Taro Watanabe$^\dag$}
  \\
  $^\dag$Nara Institute of Science and Technology, Japan
  \\
  \texttt{\{qu.zhi.pv5, taro\}@is.naist.jp}
  \\
  $^\ddagger$National Institute of Information and Communications Technology, Japan
  \\
  \texttt{\{yiran.wang, chenchen.ding, hideki.tanaka, mutiyama\}@nict.go.jp}
}

\begin{document}
\maketitle
\begin{abstract}
  Existing multilingual neural machine translation (MNMT) approaches mainly focus on improving models with the encoder-decoder architecture to translate multiple languages.
  However, decoder-only architecture has been explored less in MNMT due to its underperformance when trained on parallel data solely. 
  In this work, we attribute the issue of the decoder-only architecture to its lack of language transfer capability.
  Specifically, the decoder-only architecture is insufficient in encoding source tokens with the target language features.
  We propose dividing the decoding process into two stages so that target tokens are explicitly excluded in the first stage to implicitly boost the transfer capability across languages.
  Additionally, we impose contrastive learning on translation instructions, resulting in improved performance in zero-shot translation.
  We conduct experiments on TED-19 and OPUS-100 datasets, considering both training from scratch and fine-tuning scenarios.
  Experimental results show that, compared to the encoder-decoder architecture, our methods not only perform competitively in supervised translations but also achieve improvements of up to 3.39 BLEU, 6.99 chrF++, 3.22 BERTScore, and 4.81 COMET in zero-shot translations.\footnote{Partial works done during Zhi Qu's internship at ASTREC of NICT, Japan. Our codes are available at \url{https://github.com/zhiqu22/PhasedDecoder}.}
  % \footnote{All codes will be released for reproduction.}
  
\end{abstract}

\section{Introduction}\label{section:intro}
\begin{figure}[t]
    \centering
      \begin{subfigure}[b]{0.49\linewidth}
        \centering
        \includegraphics[width=0.98\linewidth]{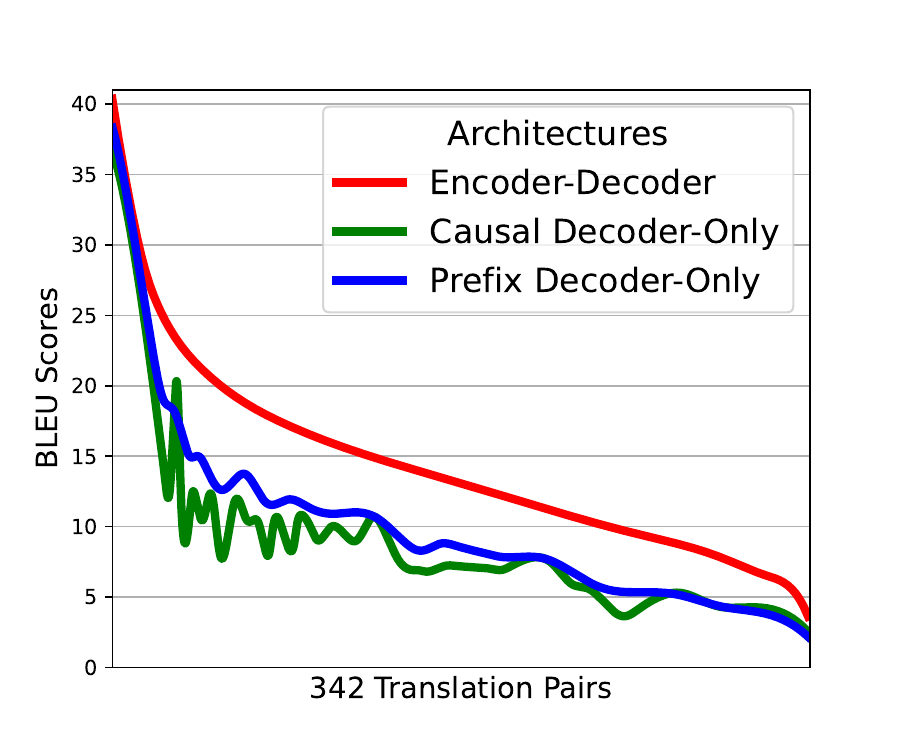}
        \caption{Performance}
        \label{fig:comparison_performance}
      \end{subfigure}
      \begin{subfigure}[b]{0.49\linewidth}
        \centering
        \includegraphics[width=\linewidth]{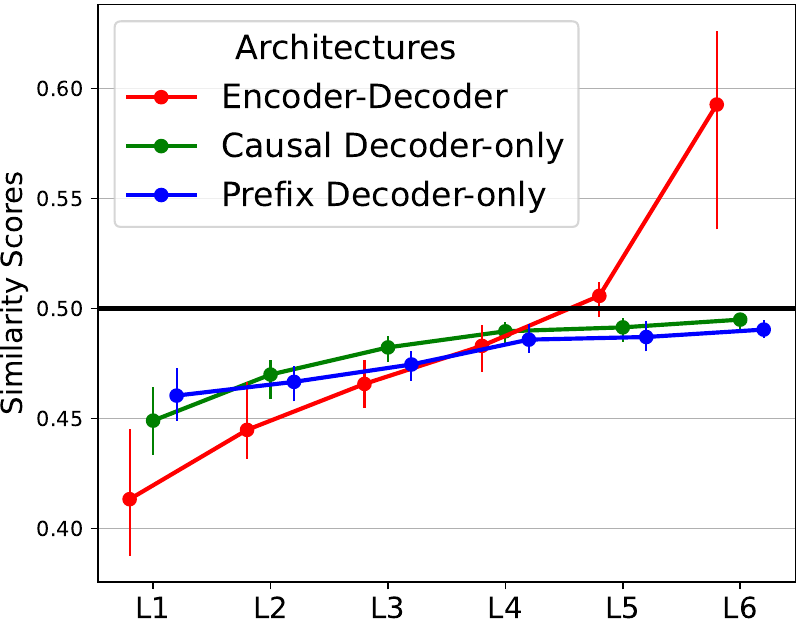}
        \caption{Language Feature}
        \label{fig:preference}
      \end{subfigure}
    \caption{Comparison between different architectures in preliminary experiments on TED-19.
    Figure \ref{fig:comparison_performance} shows the BLEU score.
    Figure \ref{fig:preference} shows the layer-wise language feature representations of a sentence where the x-axis indicates the layer number and the vertical line indicates the value range.
    Specifically, we follow \citet{transfer} to compute a similarity score, where values higher than 0.5 mean the representation exhibits the target language features more and lower than 0.5 indicates showing more source language features.
    Appendix \ref{appendix:preference} provides the details of implementation.}
    \vspace{-1.4em}
    \label{fig:comparison}
\end{figure}

Multilingual neural machine translation (MNMT) \cite{ZeroMNMT2016} aims to integrate multiple language translation directions into a single model.
Although multilingual translation systems based on large language models have demonstrated strong performance \cite{bayling, bigtrans, ftllm}, current MNMT models with the encoder-decoder architecture \cite{m2m, flores101, nllb} remain a focus of research due to the competitive performance, fewer parameters, and reduced training costs \cite{mnmtVSllm}.
However, in MNMT, models with the decoder-only architecture\footnote{The term "decoder-only architecture" encompasses both causal decoder-only architectures \cite{gpt1} and prefix decoder-only architectures \cite{unilm}.} have shown underperformance by the empirical research of \citet{decoderonlymt,decoderonly_2022}, as further evidenced by Figure \ref{fig:comparison_performance}. 
Therefore, addressing the underdevelopment of decoder-only architectures in MNMT is crucial due to the advantage of zero-shot generalization \cite{llm_zeroshot}, which potentially benefits zero-shot translation, i.e., translating language pairs unseen during training.

We attribute the issue to the lack of language transferability, causing generations to rely solely on representations that always manifest the source language features.
Specifically, MNMT encoder-decoder models typically add a language tag indicating the target language at the beginning of the source tokens as a translation instruction \cite{GooglesMNMT,TagMatter_2021}, then, \citet{investigateMNMT, transfer} show that the encoder of MNMT models transfers source tokens to represent target language features more than source language features.
As shown in Figure \ref{fig:preference}, the representation of source tokens extracted from the model with the encoder-decoder architecture mainly exhibits the target language features at the output of the encoder (red line), however, this characteristic is absent in decoder-only architectures (green and blue lines).
We hypothesize that the decoder-only architectures merely capture the surface information of source tokens instead of transferring source tokens into a state with more target language features.

We propose dividing the decoder-only architecture into two stages, namely, Two-stage Decoder-only (TDO).
Specifically, the representations of target tokens are excluded in the first stage to enforce language transfer using the translation instruction, and the target tokens are fused in the second stage, which follows the normal decoder-only manner.
Moreover, unlike the encoder-decoder architecture, where source and target tokens are processed separately, in the decoder-only architecture, source tokens pass through all layers.
However, the training objective of MNMT only focuses on the target tokens, leading to the degradation of the target language features on the source token representation.
Thus, we introduce Instruction-level Contrastive Learning (InstruCL) as a training objective to supervise source tokens in the second stage.

We evaluate the proposed methodologies on two datasets, TED-19 \cite{ted}, and OPUS-100 \cite{massive-2020, TLP-2021}, using four automatic evaluation metrics: BLEU \cite{bleu,sacrebleu}, chrF++ \cite{chrf,chrf++}, BERTScore \cite{bertscore} and COMET \cite{comet}.
Experimental results show that, compared to encoder-decoder models, our models perform competitively in supervised translations and achieve improvements of up to 3.39 BLEU, 6.99 chrF++, 3.22 BERTScore, and 4.81 COMET in zero-shot translations.
We also analyze the variation of layer-wise representations at the sentence level to demonstrate the effects of our proposed methods.
Results prove that the gains of proposed methods in the decoder-only architecture derived from improving language transfer.

\section{Related Work}
Although the large language model based on the decoder-only architecture performs satisfactorily in the multilingual translation \cite{mnmtVSllm, ftllm}, the SOTA models specialized on MNMT are still based on the encoder-decoder architecture \cite{m2m,nllb} due to the balance between costs and performances.
\citet{decoderonlymt, decoderonly_2022} empirically show that the decoder-only architecture does not have a distinct advantage in MNMT, and \citet{surveyMNMT, t5} demonstrate that the reason could be the onefold style of training data comprising only translations, degrading the zero-shot ability of the decoder-only architecture \cite{gpt3, llm_zeroshot}.

Recent investigations of the encoder-decoder architecture in MNMT reveal the deficiency of the decoder-only architecture at the representation level.
\citet{investigateMNMT, transfer-2023} point out that the sentence representations translating to two different target languages are gradually separated with the increase of layers.
\citet{transfer} demonstrate that the encoder of MNMT model transfers the source sentence representation to the target side, leading to the representation of source tokens used in the generation with more target language features.
This finding aligns with the prior empirical studies \cite{TagMatter_2021, adaptingZero, lsls-2023}, which shows that increasing target language information can lead to performance improvements. 
Moreover, this also supports our hypothesis that the weakness of the decoder-only architecture can be attributed to the lack of language transfer.

\section{Backgrounds}\label{section:background}
\subsection{Multilingual Neural Machine Translation}
A parallel multilingual corpus, denoted by $\set{C}$, consists of translation pairs in the form of $(\ve{x}, \ve{y})$.
Here, $\ve{x} = x_1, \ldots, x_I$ is the source sentence comprising $I$ tokens, and $\ve{y} = y_1, \ldots, y_J$ is the target sentence with $J$ tokens.
We also denote language tags by $\ve{l}=l_1, \ldots, l_K$, where each tag is an artificial token uniquely corresponding to one of the $K$ languages in $\set{C}$.
To serve as a translation instruction, we add the language tag specifying the target language at the beginning of the source tokens \cite{GooglesMNMT, TagMatter_2021}, denoted by $l_{\ve{y}}$.\footnote{Appendix \ref{appendix:strategy} shows the comparison between different strategies of translation instructions in MNMT.}
Thus, the training data comprises instances in the form of $(l_{\ve{y}}, \ve{x},\ve{y})$.
The model is trained over all instances in $\set{C}$ by the standard cross-entropy objective:
\vspace{-1.5em}\begin{equation}\label{eq:ce}
\mathcal{L}_{\text{ce}}= -\sum_{
  l_{
    \ve{y}}, \ve{x}, \ve{y} \in \set{C}}{\sum_{j=1}^{J}
    {
      \log~{p(y_{j} \mid l_{\ve{y}}, \ve{x}, \ve{y}_{<j})
    },
  }
}
\vspace{-0.5em}
\end{equation}
where $p(y_{j} \mid l_{\ve{y}}, \ve{x}, \ve{y}_{<j}) $ is a probability distribution for each token generated by MNMT model.
\begin{figure}[t]
    \centering
    \includegraphics[width=0.6\linewidth]{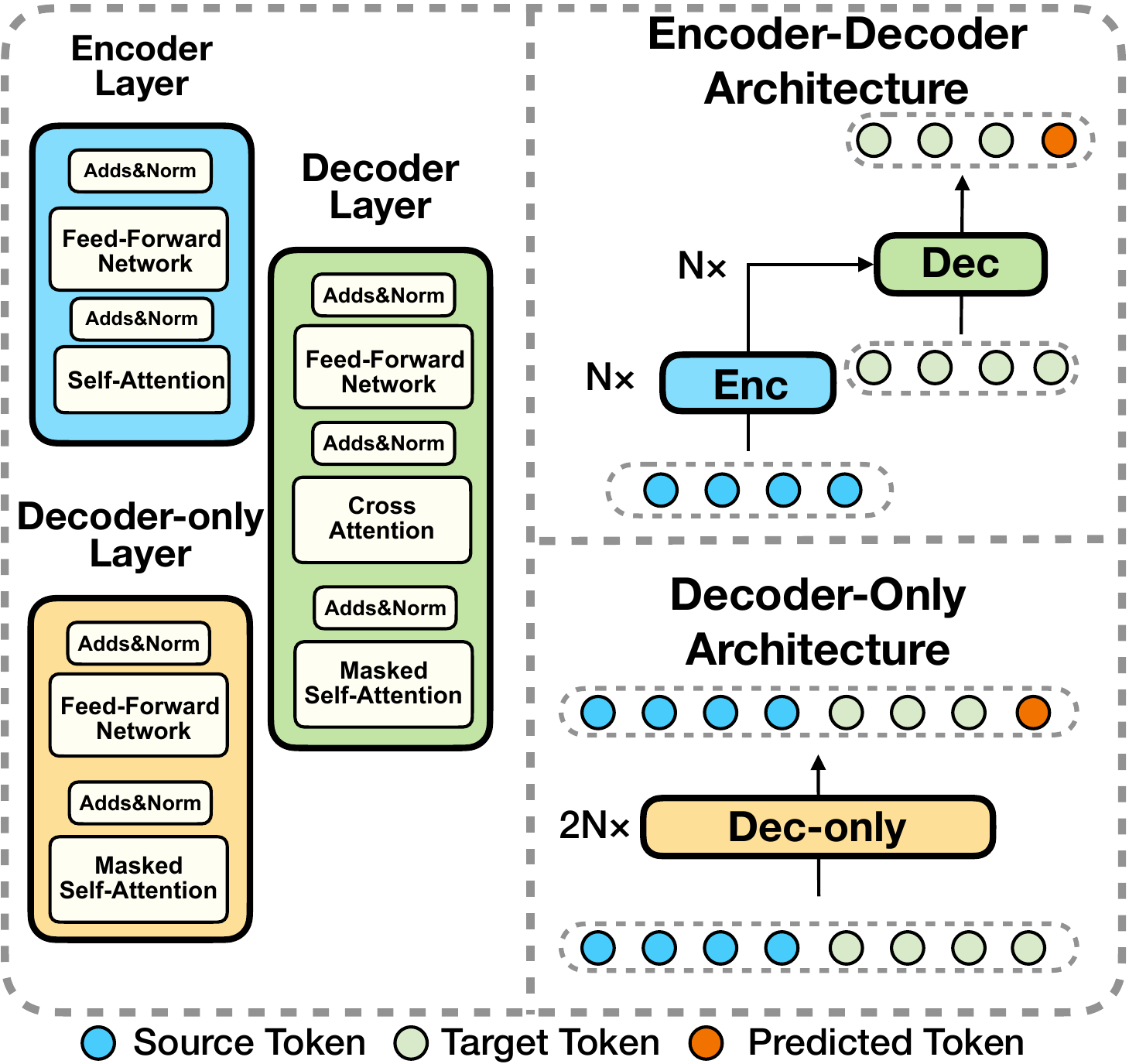}
    \vspace{-1em}
    \caption{Illustration of the encoder-decoder architecture and the decoder-only architecture.}
    \label{fig:background}
    \vspace{-1em}
\end{figure}

\vspace{-1em}
\subsection{Architectures}
\label{section:architectures}
\begin{figure*}[t]
    \centering
    \includegraphics[width=0.85\linewidth]{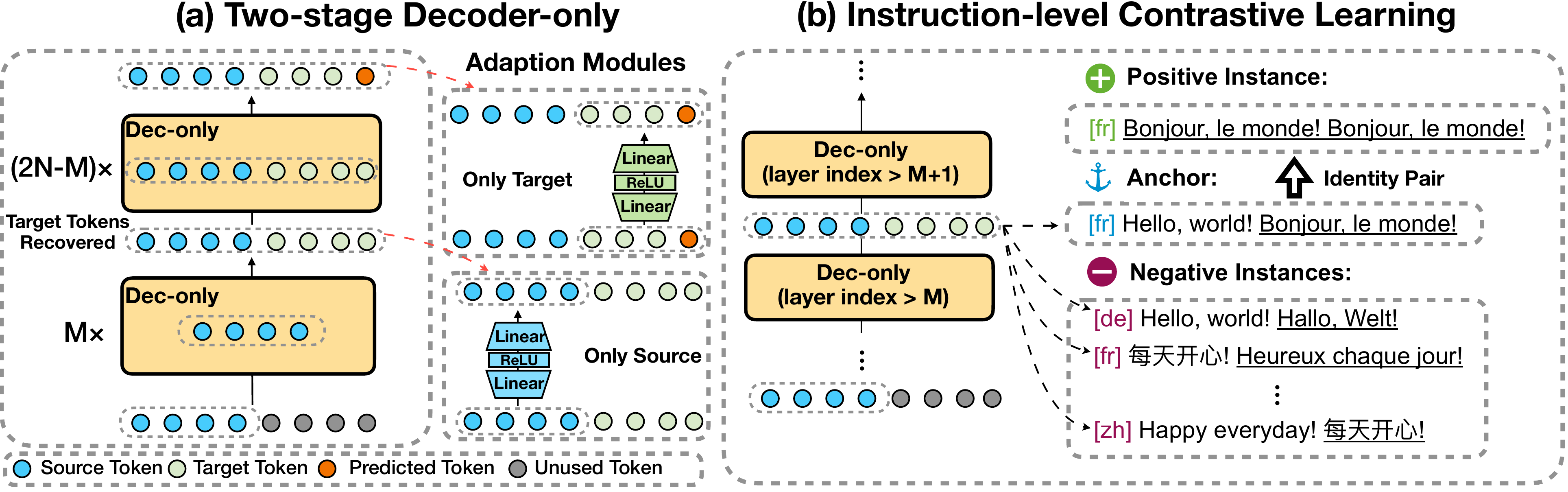}
    \vspace{-0.6em}
    \caption{Illustration of proposed methods.
    Notably, the term, Token, not only means the real token before and after the processing of model, but also refers to the representation in the corresponding position.
    (a) shows the Two-stage Decoder-only and shows the Adaption, i.e., using an additional FFN to narrow the gap between source and target representations by non-linear transformation.
    (b) shows the Instruction-level Contrastive Learning.
    Underline marks target tokens, and [*] means the instruction of this instance.
    For the anchor, negative instances in this figure meet at least one of two features: 1) different target language and 2) unparallel semantics.}
    \vspace{-1.2em}
    \label{fig:main}
\end{figure*}
All architectures discussed in this work follow the Transformer architecture \cite{transformer}, and almost all MNMT models are based on the encoder-decoder architecture \cite{GooglesMNMT, m2m, nllb, t5}, as illustrated in Figure \ref{fig:background}. It comprises an encoder and a decoder in which both are composed of $N$ layers with each encoder layer comprising a self-attention mechanism and a feed-forward network (FFN), and with each decoder layer comprising a masked self-attention mechanism, a cross-attention mechanism, and an FFN.
The encoder receives $I+1$ tokens combining by $(l_{\ve{y}}, \ve{x})$\footnote{The operation of combining means adding $l_{\ve{y}}$ at the beginning of $\ve{x}$. Appendix \ref{appendix:input} shows the specific forms in detail.}, and output the representations $\ma{H}=\seq{\ve{h}_{1},...,\ve{h}_{I+1}}, \ve{h}\in \mathbb{R}^{d}$, $d$ is the model dimension.
Then, the decoder relies on $\ma{H}$ and $\ve{y}_{<j}$ to generate the next token:
\begin{gather}\label{enc_dec}
\ma{H}^N = \operatorname{encoder}(l_{\ve{y}}, \ve{x}),\\
y_{j} = \operatorname{decoder}(\ma{H}^N, \ve{y}_{<j}),
\end{gather}
where $\ma{H}^N$ is an intermediate state used in the cross-attention mechanism in each decoder layer without further transformation.
Thus, Equation \ref{eq:ce} implicitly aligns the output of the encoder in the representational subspace of the target language, i.e., the language transfer as shown in the red line of Figure \ref{fig:preference}, because the ideal decoder should translate two sentences $\ve{x}^a$ and $\ve{x}^b$, which have the same target language, parallel semantics, and different source languages, to the same target sentence $\ve{y}$.
Formally, an ideal encoder meets the following:
\begin{equation}\label{align}
\operatorname{encoder}(l_{\ve{y}}, \ve{x}^a) = \operatorname{encoder}(l_{\ve{y}}, \ve{x}^b).
\end{equation}

A decoder-only architecture refers to a model that consists solely of a decoder (Figure \ref{fig:background}).
Each decoder-only layer consists of a masked self-attention mechanism and an FFN \cite{gpt1}, and each model has $2N$ layers to approximately match the parameter size of an encoder-decoder architecture.
We define the decoder-only process as follows: 
\vspace{-0.3em}\begin{equation}\label{dec_only}
y_{j} = \operatorname{decoder-only}(l_{\ve{y}} , \ve{x}, \ve{y}_{<j}).
\end{equation}\vspace{-0.3em}
Notably, the difference between $\operatorname{decoder-only}(\cdot)$ and $\operatorname{decoder}(\cdot)$ is that $\operatorname{decoder-only}(\cdot)$ fuses the source and target information by a concatenated input, namely, $l_{\ve{y}}, \ve{x}$, and $\ve{y}$ are equally treated\footnote{Appendix \ref{appendix:input} compares the difference of the input and output forms between encoder-decoder and decoder-only models.}, instead of using a cross-attention mechanism.
Thus, there exists no intermediate state to align different source languages as Equation \ref{align}, resulting in the blue and green lines of Figure \ref{fig:preference}.
Moreover, we follow \citet{decoderonlymt,t5} to distinguish the decoder-only by the manner of masked self-attention mechanism as causal decoder-only and prefix decoder-only (Appendix \ref{appendix:attention}).
Finally, compared to the encoder-decoder architecture, the decoder-only architecture requires around 10\% fewer parameters (Appendix \ref{appendix:estimation}).

\section{Methodologies}\label{section:methodologies}
\subsection{Two-stage Decoder-only Architecture}\label{section:tdo}
The limitations of the decoder-only architecture in MNMT likely arise from inadequate language transfer capabilities, i.e., the absence of Equation~\ref{align}.
To address this issue, we propose the Two-stage Decoder-only (TDO) architecture, which divides the decoder-only process into two stages to implicitly align representations of different source languages in the subspace of the target language.
Specifically, as illustrated in Figure \ref{fig:main}, the target tokens are explicitly excluded in the first stage, i.e., the first $M$ layers, and these target tokens are fused in the second stage, i.e., the subsequent $2N-M$ layers.
The process of TDO is formally expressed: 
\vspace{-0.5em}
\begin{gather}\label{tdo}
\ma{H}^M = \operatorname{decoder-only}_{1}(l_{\ve{y}}, \ve{x}),\\
\label{tdo2}
y_{j} = \operatorname{decoder-only}_{2}(\ma{H}^M, \ve{y}_{<j}),
\end{gather}
where $\operatorname{decoder-only}_{1}(\cdot)$ enables the implicit alignment as done in Equation \ref{align}.
Notably, the first stage logically acts as an encoder when prefixed masking is applied to the self-attention mechanism.
However, the first and second stages remain unified structures, and the fusing of source and target information follows the manner of $\operatorname{decoder-only}(\cdot)$ rather than $\operatorname{decoder}(\cdot)$.
Therefore, TDO architecture preserves the decoder-only architecture.

Notably, a representational gap arises at the $M+1$ layer due to our imbalance design where the source tokens have passed through the preceding $M$ layers, while the target tokens are not.
To bridge this gap, as shown in Figure \ref{fig:main}, we employ an additional FFN as an adaption module\footnote{Adaptation module is shared for all languages instead of a language-specific component \cite{bapna-2019}.} at the output of the $M$ layer to nonlinearly transform the representation of source tokens.
Similarly, since the source and target tokens share the same representational space in the second stage, we employ another adapter at the output of the $2N$ layer to ensure that the output representation of target tokens remains unaffected by the source language.

\subsection{Instruction-level Contrastive Learning}\label{section:instrucl}
Although Equation \ref{tdo} transfers $\ma{H}$, i.e., the representation of source tokens, to $\ma{H}^M$, which aligns with the target language, $\ma{H}$ potentially tends to degrade towards the source language in Equation \ref{tdo2} because Equation~\ref{eq:ce} does not supervise $\ma{H}$ directly.\footnote{Although the language modeling loss \cite{gpt1} can provide supervision for the representation of source tokens, \citet{decoderonlymt} show that supervising the representation of source tokens does not benefit MNMT.}

Contrastive learning, which is a technique to encourage representations towards the target states \cite{contrastive_survey}, is helpful to mitigate this degradation.
However, two challenges remain in this process.
The first is the lack of optimization objectives for aligning $\ma{H}$ with the target language.
For instance, the $\ma{H}$ derived by a translation from German to English cannot be considered an anchor to optimize another $\ma{H}$ derived by a translation from French to English because neither adequately represents the optimal state of English.
The second challenge is that the optimization at the sentence representation level potentially leads to suboptimal results.
For instance, \citet{Constras-2021} suggest averaging representations of all tokens to get a sentence representation for contrastive learning, which loses the syntactic information. 

We propose Instruction-level Contrastive Learning (InstruCL), which only aligns $l_{\ve{y}}$, i.e., the translation instruction, of each instance, given that MNMT remains sensitive to $l_{\ve{y}}$ \cite{TagMatter_2021}.
As shown in Figure \ref{fig:main}, given an anchor $(l_{\ve{y}}, \ve{x},\ve{y})$, we establish an identity pair in the form of $(l_{\ve{y}}, \ve{y},\ve{y})$, namely a pseudo pair translating the target sentence to itself, as the positive instance because the identity pair can serve as a proxy for the target language \cite{transfer}.
Specifically, in a training batch, we have a set of representations $ \set{B} = \seq{\ve{h}^{1}_1, \ve{h}^{2}_1, \dots}$ where $\ve{h}_{1}$ is the representation of $l_{\ve{y}}$ collected from $\ma{H}$.
Then, we designate one instance of $ \set{B} $ as the anchor, denoted by $ \ve{h}^{\text{anc}} $.
Other instances are treated as negative instances, which meet one or both of the following features compared to the anchor: different target languages or unparallel semantics.
Subsequently, the identity pair established by the anchor would be fed into the model and we collect the representation of $l_{\ve{y}}$ at the same layer, and denote it by $ \ve{h}^{\text{pos}} $.
The objective of InstruCL is formulated as:
\vspace{-0.5em}\begin{equation}
\label{cl}
\begin{split}
    \mathcal{L}_{\text{ctr}} &= -
    \sum_{\ve{h} \in \set{B}}
    \log
    \frac
    {\operatorname{exp}({s^+})}
    {\operatorname{exp}({s^+}) + \sum_{i=1}^{|\set{B}|-1} \operatorname{exp}({s_i^-})}, \\
    s^+ &= \operatorname{sim}(\ve{h}^{\text{anc}}, \ve{h}^{\text{pos}}), \\
    s_i^- &= \operatorname{sim}(\ve{h}^{\text{anc}}, \ve{h}^i_1), \ve{h}^i_1 \neq \ve{h}^{\text{anc}},
\end{split}
\end{equation}
where $\operatorname{sim}(\cdot)$ calculates the similarity of representations using the cosine similarity.
The final training objective is simply jointed as:
\begin{equation}
\label{loss}
    \mathcal{L} = \mathcal{L}_{\text{ce}} + \mathcal{L}_{\text{ctr}}.
\end{equation}

\section{Experiments}\label{section:experiments}
\subsection{Datasets and Evaluations}\label{section:dataset}
Following prior works \cite{TagMatter_2021, decoderonly_2022,  zero-2023, transfer-2023, transfer}, we use English-centric datasets in our experiments, where the training and validation data consist of translation pairs both from English and to English.
It is an ideal setup for the evaluation of zero-shot translation capabilities, because non-central languages have never seen each other.
We utilize two datasets in our experiments: 1) TED-19 \cite{transfer}, a sub-collection of TED Talks \cite{ted}, comprising 6.5 million instances across 19 languages from various language families; and 2) OPUS-100 \cite{massive-2020, TLP-2021}, which includes 95 languages and a total of 92 million instances. Detailed information about these datasets is provided in Appendix \ref{appendix:dataset}.

We set the beam size to 4 during inference and evaluate the output quality using four automatic evaluation metrics for a comprehensive assessment: SacreBLEU \cite{bleu, sacrebleu}, chrF++ \cite{chrf, chrf++}, BERTScore \cite{bertscore}, and COMET \cite{comet}. Additionally, we employ \textit{fasttext-langdetect}\footnote{\url{https://pypi.org/project/fasttext-langdetect}} to measure the target-off ratio on zero-shot pairs, i.e., the ratio of cases where the source sentence is not translated into the correct target language, as a secondary metric.
Our selection criteria for these evaluation metrics are further described in Appendix \ref{appendix:standards}.

\subsection{Experimental Setups}\label{section:setup}
We conduct experiments from two perspectives: training from scratch and fine-tuning.
Based on the findings by \citet{decoderonlymt,decoderonly_2022}, which empirically demonstrate that the decoder-only architecture underperforms compared to the encoder-decoder architecture in MNMT, and our motivation, which aims to improve the decoder-only architecture, our baselines are vanilla models with the encoder-decoder and decoder-only architectures.
Specifically, we train models with the encoder-decoder architecture from scratch using TED-19 and OPUS-100 as baselines.
Additionally, we fine-tune three pre-trained models with the encoder-decoder architecture, namely M2M-418M \cite{m2m}, NLLB-600M \cite{nllb}, and M2M-1.2B \cite{m2m}, using TED-19 as baselines.
Moreover, although the proposed methods are not restricted to a specific architecture, the adaptation modules are not implemented for the models with the encoder-decoder architecture, because, when the hyper-parameters are consistent, the decoder-only architecture with adaptation modules still contains fewer learnable parameters\footnote{Appendix \ref{appendix:model} lists the count of modeling parameters for different cases in detail.} to ensure fairness, i.e., models have the same magnitude of parameters.
We further conduct experiments that apply InstruCL to models with different architectures.
The experimental results and discussions are provided in Appendix \ref{appendix:effect} as assisted evidence to support the motivation in Section \ref{section:instrucl}, namely, InstruCL supplements the inadequate supervision of Equation \ref{eq:ce} in the second stage.

Our models in this work conform to the manner of the Transformer \cite{transformer}.
For training from scratch, we configure the models with $N=6$, $d=512$, and an FFN inner size of $4d$ for models trained on TED-19.
The FFN in the adaptation module is dimensionally matched to the FFN in the main network. For OPUS-100, we explore both a deeper model with $N=12$ and a wider model with $N=6$ and $d=1024$.
Fine-tuning experiments are conducted solely on TED-19.
Given that pre-trained models for MNMT typically employ an encoder-decoder architecture, we initialize our model's parameters from the decoder, freezing the embedding layer during training.
For M2M-418M and NLLB-600M, we set $N=6$, and for M2M-1.2B, we set $N=12$, maintaining the original settings for $d$ and the FFN inner size.
To ensure comparability across different architectures, we consistently set $M=N$ and the layer index of InstruCL to $1.5N$ in the main experiments.
Detailed settings for training and the count of learnable parameters can be found in Appendix \ref{appendix:model}.

\begin{table*}[!ht]
    \centering
    \resizebox{\textwidth}{!}{
    \begin{tabular}{l ccccccccccccccccc}
    \toprule
        ~ & ~ & ~ & ~ & ~ & \multicolumn{3}{c}{BLEU $\uparrow$ } & \multicolumn{3}{c}{chrF++ $\uparrow$ } & \multicolumn{3}{c}{BERTScore $\uparrow$ }&\multicolumn{3}{c}{COMET $\uparrow$ }&off $\downarrow$ \\ 
    \midrule
        ~ & ~ & Pref. & Adap. & CL & \texttt{en}$\to$ & $\to$\texttt{en} & zero & \texttt{en}$\to$ & $\to$\texttt{en} & zero & \texttt{en}$\to$ & $\to$\texttt{en} & zero & \texttt{en}$\to$ & $\to$\texttt{en} & zero &zero\\ 
    \midrule
        \multirow{11}{*}{\begin{minipage}{0.9cm} TED\\ N$=$6 \\$d=$512 \end{minipage}} & Enc-Dec & ~ & ~ & ~ & 25.46  & 28.31  & 12.32  & 45.96  & 50.86  & 32.13  & 84.10  & 93.37  & 78.03  & 80.49  & 78.15  & 67.26  &3.82\\ 
        \cdashline{3-18}
        ~ & \multirow{2}{*}{Dec-only} & ~ & ~ & ~ & 22.54  & 24.14  & 7.33  & 42.84  & 45.08  & 23.36  & 82.96  & 92.31  & 74.38  & 76.60  & 72.99  & 57.50  &6.01\\ 
        ~ & ~ & \checkmark & ~ & ~ & 24.00  & 26.97  & 8.18  & 44.49  & 48.93  & 25.35  & 83.54  & 92.97  & 74.52  & 78.46  & 76.10  & 56.74  &5.51\\ 
        \cdashline{3-18}
        ~ & \multirow{8}{*}{TDO} & ~ & ~ & ~ & 25.47  & 28.88  & 13.56  & 45.98  & 51.33  & 34.04  & 84.11  & 93.45  & 78.90  & 80.41  & 78.42  & 69.74  & 3.54\\ 
        ~ & ~ & ~ & \checkmark & ~ & 25.55  & \textbf{28.98}  & 13.61  & 46.03  & \textbf{51.49}  & 34.11  & 84.15  & 93.50  & 78.94  & 80.56  & \textbf{78.65}  & 70.09 & 3.49 \\ 
        ~ & ~ & ~ & ~ & \checkmark & 25.37  & 28.46  & 13.95  & 45.99  & 51.13  & 34.41  & 84.09  & 93.40  & 79.15  & 80.35  & 78.26  & 70.43  &3.45\\ 
        ~ & ~ & ~ & \checkmark & \checkmark & 25.60  & 28.82  & 14.16  & 46.11  & 51.35  & 34.76  & 84.13  & 93.45  & 79.29  & 80.52  & 78.47  & 70.98 &3.43 \\ 
        \cdashline{3-18}
        ~ & ~ & \checkmark & ~ & ~ & 25.53  & 28.76  & 14.26  & 46.01  & 51.09  & 34.72  & 84.13  & 93.41  & 79.27  & 80.43  & 78.18  & 70.82  & 3.43\\ 
        ~ & ~ & \checkmark & \checkmark & ~ & 25.61  & 28.52  & 14.51  & 46.04  & 50.89  & 35.01  & \textbf{84.16}  & 93.40  & 79.41  & 80.60  & 78.16  & 71.48 &3.49 \\ 
        ~ & ~ & \checkmark & ~ & \checkmark & \textbf{25.62}  & 28.94  & 14.70  & \textbf{46.15}  & 51.46  & 35.34  & 84.15  & \textbf{93.47}  & 79.57  & 80.55  & 78.55  & 71.94 & \textbf{3.39}  \\ 
        ~ & ~ & \checkmark & \checkmark & \checkmark & 25.61  & 28.66  & \textbf{14.81}  & 46.05  & 51.01  & \textbf{35.35}  & \textbf{84.16}  & 93.41  & \textbf{79.60}  & \textbf{80.61}  & 78.22  & \textbf{72.07} & 3.42 \\ 
    \midrule
        \multirow{7}{*}{\begin{minipage}{0.9cm} OPUS\\ N$=$12 \\$d=$512 \end{minipage}} & Enc-Dec & ~ & ~ & ~ & \textbf{25.18}  & 29.79  & 5.13  & \textbf{44.75}  & 48.40  & 12.95  & \textbf{82.98}  & 92.33  & 72.44  & \textbf{76.59}  & 76.21  & 58.51  & 64.21\\ 
        \cdashline{3-18}
        ~ & \multirow{2}{*}{Dec-only} & ~ & ~ & ~ & 23.09  & 26.80  & 5.42  & 42.18  & 45.05  & 13.55  & 82.19  & 91.72  & 72.48  & 74.66  & 73.65  & 58.17  &60.22 \\ 
        ~ & ~ & \checkmark & ~ & ~ & 23.96  & 28.41  & 6.62  & 42.98  & 47.22  & 15.36  & 82.47  & 92.06  & 73.57  & 75.48  & 75.34  & \textbf{59.56}  &58.91 \\ 
        \cdashline{3-18}
        ~ & \multirow{4}{*}{TDO} & \checkmark & ~ & ~ & 24.88  & \textbf{29.97}  & 5.32  & 44.72  & \textbf{49.39}  & 13.29  & 82.91  & \textbf{92.41}  & 72.50  & 76.26  & \textbf{76.73}  & 58.30 & 51.56\\ 
        ~ & ~ & \checkmark & \checkmark & ~ & 24.79  & 29.22  & 5.97  & 44.69  & 48.35  & 14.30  & 82.87  & 92.34  & 72.97  & 76.04  & 76.25  & 58.33  & 53.80 \\ 
        ~ & ~ & \checkmark & ~ & \checkmark & 24.35  & 29.52  & 7.93  & 44.44  & 48.74  & 18.65  & 82.84  & 92.37  & 73.97  & 75.93  & 76.23  & 58.71  & 48.37 \\ 
        ~ & ~ & \checkmark & \checkmark & \checkmark & 24.73  & 29.70  & \textbf{8.52}  & 44.60  & 48.72  & \textbf{19.94}  & 82.90  & 92.38  & \textbf{74.32}  & 76.16  & 76.59  & 58.82 & \textbf{43.38} \\ 
    \midrule
        \multirow{7}{*}{\begin{minipage}{0.9cm} OPUS\\ N$=$6 \\$d=$1024 \end{minipage}} & Enc-Dec & ~ & ~ & ~ & \textbf{27.71}  & 31.60  & 6.95  & 46.84  & 50.31  & 15.89  & 83.55  & 92.62  & 74.12  & \textbf{78.10}  & 77.58  & 59.99 & 57.15\\ 
        \cdashline{3-18}
        ~ & \multirow{2}{*}{Dec-only} & ~ & ~ & ~ & 26.09  & 29.09  & 7.55  & 44.51  & 47.44  & 16.98  & 82.93  & 92.12  & 73.94  & 76.77  & 75.80  & 61.21  & 63.80 \\ 
        ~ & ~ & \checkmark & ~ & ~ & 26.79  & 30.42  & 8.15  & 45.48  & 48.92  & 17.65  & 83.21  & 92.37  & 74.17  & 77.53  & 76.69  & \textbf{62.32} & 55.67  \\ 
        \cdashline{3-18}
        ~ & \multirow{4}{*}{TDO} & \checkmark & ~ & ~ & 27.22  & 31.58  & 7.06  & 46.54  & \textbf{50.59}  & 15.96  & 83.44  & 92.64  & 73.78  & 77.68  & \textbf{77.89}  & 60.60  & 52.43\\ 
        ~ & ~ & \checkmark & \checkmark & ~ & 27.51  & \textbf{31.64}  & 7.70  & \textbf{46.87}  & 50.39  & 17.32  & \textbf{83.58}  & 92.58  & 74.32  & 78.05  & 77.58  & 61.24  & 49.87\\ 
        ~ & ~ & \checkmark & ~ & \checkmark & 27.12  & 31.49  & 9.28  & 46.55  & 50.23  & 21.33  & 83.50  & \textbf{92.65}  & \textbf{75.04}  & 77.63  & 77.64  & 60.84  & \textbf{39.71} \\ 
        ~ & ~ & \checkmark & \checkmark & \checkmark & 27.45  & 31.36  & \textbf{9.36}  & 46.79  & 50.06  & \textbf{21.05}  & 83.52  & 92.64  & 74.88  & 77.97  & 77.75  & 61.78  & 43.36\\ 
    \bottomrule
    \end{tabular}}
    \caption{Averaged scores of results in the experiments of training from scratch.
    Enc-Dec and Dec-only are abbreviations of encoder-decoder and decoder-only, respectively.
    Pref., Adap., and Cl abbreviates Prefix, Adaption and InstruCL, respectively.
    \checkmark in the Prefix column means the masked self-attention mechanism follows Prefix manner, conversely, follows Causal manner.
    \texttt{en}$\to$ and $\to$\texttt{en} means the supervised pairs translating from English to non-central languages and translating from non-central languages to English, respectively.
    zero abbreviates zero-shot pairs, off abbreviates the target-off ratio.
    The best score in each column and block is in bold.}
    \vspace{-1em}
  \label{tab:result1}
\end{table*}

\subsection{Results: Training from scratch}\label{section:result1}
Table \ref{tab:result1} shows the experimental results.
The comparison between the basic architectures shows that, first, the prefix decoder-only consistently outperforms the causal decoder-only, which aligns with \citet{t5}.
Second, the decoder-only architecture consistently underperforms the encoder-decoder architecture in supervised pairs of all three settings, with maximum deficits of -4.17, -5.78, -1.14, and -5.16 on the BLEU, chrF++, BERTScore, and COMET respectively.
On the other hand, while the decoder-only architecture shows weaker performance on TED-19 for zero-shot translation, it achieves higher scores in two settings on OPUS-100.
This suggests that the zero-shot capability of the decoder-only architecture in MNMT relates to the amount of data and parameters.

In comparison with the encoder-decoder architecture, TDO, firstly, achieves competitively supervised capabilities using fewer parameters, and, specifically, TDO is slightly stronger when translating to \texttt{en} and slightly weaker when translating from \texttt{en}.
Secondly, our method exhibits stronger zero-shot translation scores, achieving scores improvements of +2.49, +3.22, +1.57, and +4.81; +3.39, +6.99, +1.88, and +0.31; +2.41, +5.16, +0.76, +1.79 across three settings for the four main metrics respectively.
We also find that the Adaptation module enhances both supervised and zero-shot translation performance.\footnote{Appendix \ref{appendix:params} shows the improvement is not because of increased parameters.}
On the other hand, InstruCL significantly boosts zero-shot capability, though there is a degradation in supervised translation performance.
Additionally, with the Adaptation module implemented, the degree of degradation in supervised performance is reduced.

Moreover, the prefix decoder-only architecture achieves the highest COMET score on OPUS-100, though, it remains weaker on BERTScore compared to TDO, where both two metrics are based on semantics.
This phenomenon can be explained by the target-off ratio, in which models with decoder-only architecture still have a high target-off ratio with biasing towards English primarily \cite{off} to hamper the evaluation of COMET by considering the source sentence at the same time.

\begin{table*}[!ht]
    \centering
    \resizebox{0.82\textwidth}{!}{
    \begin{tabular}{llccccccccccccc}
    \toprule
        ~ & ~ & \multicolumn{3}{c}{BLEU $\uparrow$ } & \multicolumn{3}{c}{chrF++ $\uparrow$ } & \multicolumn{3}{c}{BERTScore $\uparrow$ }&\multicolumn{3}{c}{COMET $\uparrow$ } &off $\downarrow$\\ 
    \midrule
        ~ & ~ & \texttt{en}$\to$ & $\to$\texttt{en} & zero & \texttt{en}$\to$ & $\to$\texttt{en} & zero & \texttt{en}$\to$ & $\to$\texttt{en} & zero & \texttt{en}$\to$ & $\to$\texttt{en} & zero& zero \\ 
    \midrule
        \multirow{6}{*}{\begin{minipage}{0.7cm} M2M\\ 418M \end{minipage}} & Enc-Dec & 26.59  & 31.62  & 15.73  & 46.79  & 54.07  & 36.25  & 84.48  & 94.02  & 80.12  & 82.39  & 81.30  & 75.11&\textbf{3.24}  \\ 
        ~ & Dec-only & 25.72  & 30.06  & 14.67  & 45.88  & 52.52  & 34.51  & 84.12  & 93.70  & 79.45  & 81.61  & 79.89  & 73.33 &3.51 \\ 
        ~ & TDO & 26.63  & \textbf{32.44}  & 15.96  & 46.90  & 54.80  & 36.56  & 84.49  & 94.15  & 80.28  & 82.31  & 81.80  & 75.45 &\textbf{3.24}  \\ 
        ~ & \hspace{0.4em}+Adap. & \textbf{26.87}  & 31.93  & 16.12  & \textbf{47.08}  & 54.21  & 36.73  & \textbf{84.58}  & 94.08  & 80.35  & \textbf{82.62}  & 81.54  & 75.80 &3.31 \\ 
        ~ & \hspace{0.4em}+CL & 26.61  & 32.34  & 16.01  & 47.03  & \textbf{55.07}  & \textbf{36.87}  & 84.51  & \textbf{94.16}  & 80.37  & 82.29  & \textbf{81.82}  & 75.70 &3.31 \\ 
        ~ & \hspace{0.4em}+Adap.,+CL & 26.75  & 31.83  & \textbf{16.20}  & 46.98  & 54.09  & 36.82  & 84.56  & 94.07  & \textbf{80.41}  & 82.56  & 81.52  & \textbf{75.95} &3.30 \\ 
    \midrule
        \multirow{6}{*}{\begin{minipage}{0.7cm} NLLB\\ 600M \end{minipage}} & Enc-Dec & 26.39  & 32.04  & 15.44  & 46.90  & 54.51  & 36.09  & 84.46  & 94.07  & 79.96  & 81.98  & 81.16  & 74.05 & 3.42 \\ 
        ~ & Dec-only & 26.35  & 30.20  & 14.69  & 46.36  & 51.96  & 34.16  & 84.35  & 93.72  & 79.45  & \textbf{82.20}  & 79.94  & 73.62 & 3.63 \\ 
        ~ & TDO & 25.82  & 32.15  & 15.48  & 46.42  & 54.76  & 36.35  & 84.30  & 94.10  & 80.09  & 81.34  & 81.28  & 74.17 & \textbf{3.28} \\ 
        ~ & \hspace{0.4em}+Adap. & \textbf{26.60}  & \textbf{32.47}  & 15.82  & \textbf{47.04}  & \textbf{54.83}  & \textbf{36.62}  & \textbf{84.54}  & \textbf{94.15}  & 80.23  & 82.08  & \textbf{81.48}  & 74.89 & 3.41 \\ 
        ~ & \hspace{0.4em}+CL & 25.87  & 32.29  & 15.48  & 46.44  & 54.71  & 36.21  & 84.31  & 94.11  & 80.09  & 81.43  & 81.27  & 74.18 & 3.47 \\ 
        ~ & \hspace{0.4em}+Adap.,+CL & 26.58  & 32.37  & \textbf{15.85}  & 46.94  & 54.69  & 36.52  & 84.52  & 94.14  & \textbf{80.24}  & 82.12  & 81.44  & \textbf{74.93}  & 3.36\\ 
    \midrule
        \multirow{6}{*}{\begin{minipage}{0.7cm} M2M\\ 1.2B \end{minipage}} & Enc-Dec & 27.02  & 31.75  & 16.21  & 47.05  & 53.82  & 36.51  & 84.60  & 94.03  & 80.29  & 82.93  & 81.38  & 76.13  &\textbf{3.20}\\ 
        ~ & Dec-only & 26.47  & 29.99  & 15.40  & 46.47  & 52.01  & 35.10  & 84.36  & 93.72  & 79.83  & 82.51  & 80.21  & 75.33 & 3.46 \\ 
        ~ & TDO & 27.17  & \textbf{31.95}  & 16.45  & 47.37  & \textbf{54.66}  & 37.24  & 84.64  & \textbf{94.11}  & 80.48  & 82.96  & 81.71  & 76.47 & 3.29 \\ 
        ~ & \hspace{0.4em}+Adap. & 27.32  & 31.05  & 16.57  & \textbf{47.53}  & 53.76  & \textbf{37.47}  & 84.68  & 93.99  & \textbf{80.56}  & 83.11  & 81.29  & 76.72 & 3.31 \\ 
        ~ & \hspace{0.4em}+CL & 27.27  & 31.83  & 16.57  & 47.32  & 54.42  & 37.08  & 84.67  & \textbf{94.11}  & 80.54  & 83.04  & \textbf{81.75}  & 76.72 & 3.32 \\ 
        ~ & \hspace{0.4em}+Adap.,+CL & \textbf{27.41}  & 30.72  & \textbf{16.60}  & 47.49  & 53.38  & 37.23  & \textbf{84.70}  & 93.96  & 80.55  & \textbf{83.24}  & 81.21  & \textbf{76.88} & 3.28 \\ 
    \bottomrule
    \end{tabular}}
    \vspace{-0.5em}
    \caption{Averaged scores of results in the experiments of fine-tuning. Abbreviations align with Table \ref{tab:result2}. Notably, the decoder-only and TDO architectures use Prefix masked self-attention only. The best score is in bold.}
    \vspace{-1em}
  \label{tab:result2}
\end{table*}
\subsection{Results: Fine-tuning}\label{section:result2}
Table \ref{tab:result2} shows the experimental results by fine-tuning the pre-trained models, which shows a similar tendency to Table \ref{tab:result1} in general.
First, since we initialize the model using parameters from the decoder, the training processes for the encoder-decoder, decoder-only, and TDO architectures are relatively fair.
Thus, we can conclude that, when compared with the decoder-only architecture, the proposed TDO architecture supports an efficient transformation from pre-trained encoder-decoder models.
Secondly, when compared with the encoder-decoder models, TDO models achieve the highest scores across four metrics, reaching up to +0.39, +0.48, +0.10, and +0.31 for pairs translating to \text{en}, up to +0.82, +1.00, +0.14, and +0.52 for pairs translating from \text{en}, and up to +0.47, +0.96, +0.29, and +0.88 for zero-shot pairs.
TDO models also show an improvement in the off-target ratio compared to the decoder-only models and keep the same level as the encoder-decoder models.
Moreover, we observe that InstruCL does not show significant improvements in the case of NLLB-600M, whereas it remains effective in the two M2M cases.
This may be attributed to that NLLB supports 205 languages, compared to 100 languages of M2M, implying a denser representational space that affects the effectiveness of InstruCL in aligning representations across languages.

\section{Discussion}
\subsection{Representation Analysis}
\begin{figure}[t]
    \centering
    \includegraphics[width=\linewidth]{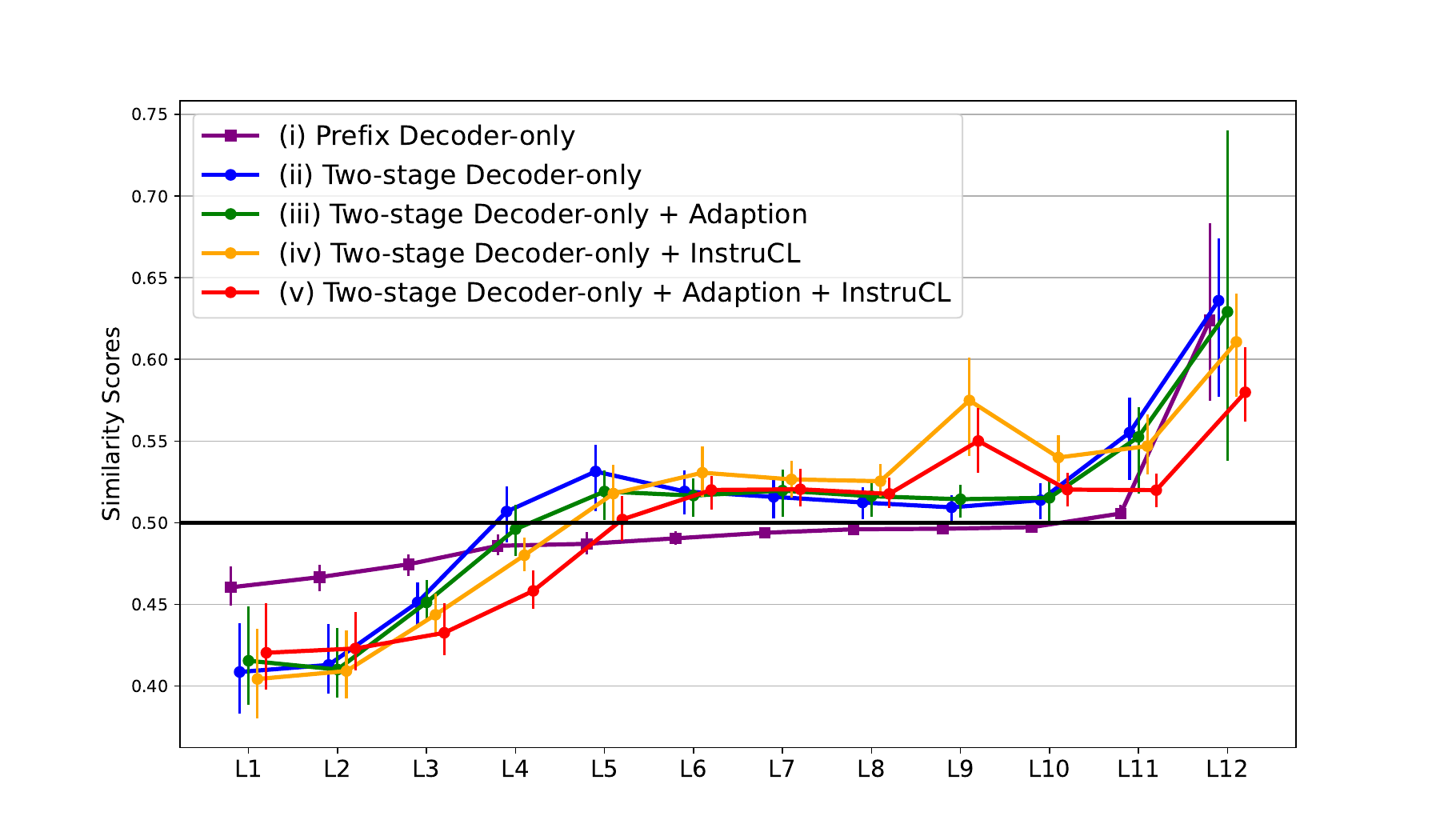}
    \caption{Illustration of linguistic preference, which follows Figure \ref{fig:preference}.
    All cases in this figure use the Prefix manner for the masked self-attention mechanism.
    The marker of prefix decoder-only is square, and our proposed methods are round.
    The x-axis is the index of layers, and the vertical line indicates the value range.
    }
    \label{fig:representation}
    \vspace{-1em}
\end{figure}
The limitation of the decoder-only architecture in MNMT is due to the lack of language transfer, which is shown in Figure \ref{fig:preference}.
To verify whether our proposed methods can address this issue, we analyze the layer-wise sentence representations of five models trained on TED-19: (\romannumeral1) a prefix decoder-only model with $N=6$; (\romannumeral2) a TDO model with $M=6$; (\romannumeral3) a TDO model with Adaption modules; (\romannumeral4) a TDO model with InstrucCL; (\romannumeral5) a TDO model with Adaption modules and InstrucCL.

As illustrated in Figure \ref{fig:representation}, the representation of (\romannumeral1) only exhibits a preference for the target language in the last two layers.
However, (\romannumeral2) shows a preference for the target language from the fourth layer, and this trend continues into the second stage.
Although (\romannumeral3) exhibits a more stable layer-wise trend compared to (\romannumeral2), it shows significant differences in the final output across languages.
Meanwhile, (\romannumeral4) exhibits smaller differences across languages.
Finally, (\romannumeral5) incorporates all the advantages of (\romannumeral3) and (\romannumeral4).
Therefore, we can conclude that the TDO enables better language transfer by aligning different languages in the representational subspace of the target language.
Meanwhile, the Adaption module and InstrucCL improve the transferability of multilingual representations.

\subsection{How to balance two stages?}\label{section:balance}
\begin{figure}[t]
    \centering
      \begin{subfigure}[b]{0.49\linewidth}
        \centering
        \includegraphics[width=\linewidth]{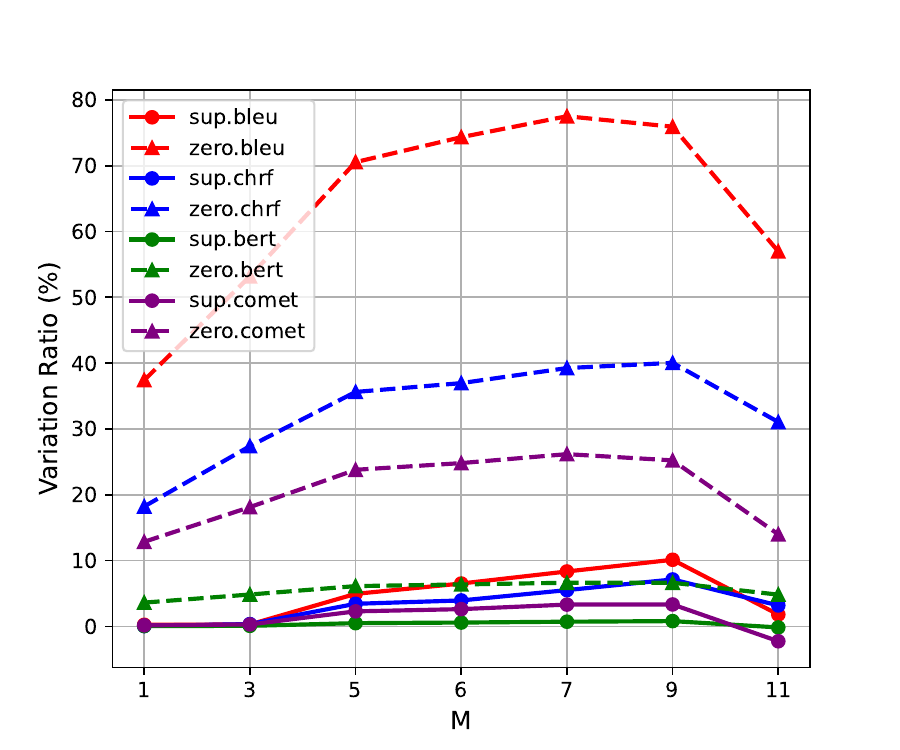}
        \caption{TED-19}
        \label{fig:split_ted}
      \end{subfigure}
      \begin{subfigure}[b]{0.49\linewidth}
        \centering
        \includegraphics[width=\linewidth]{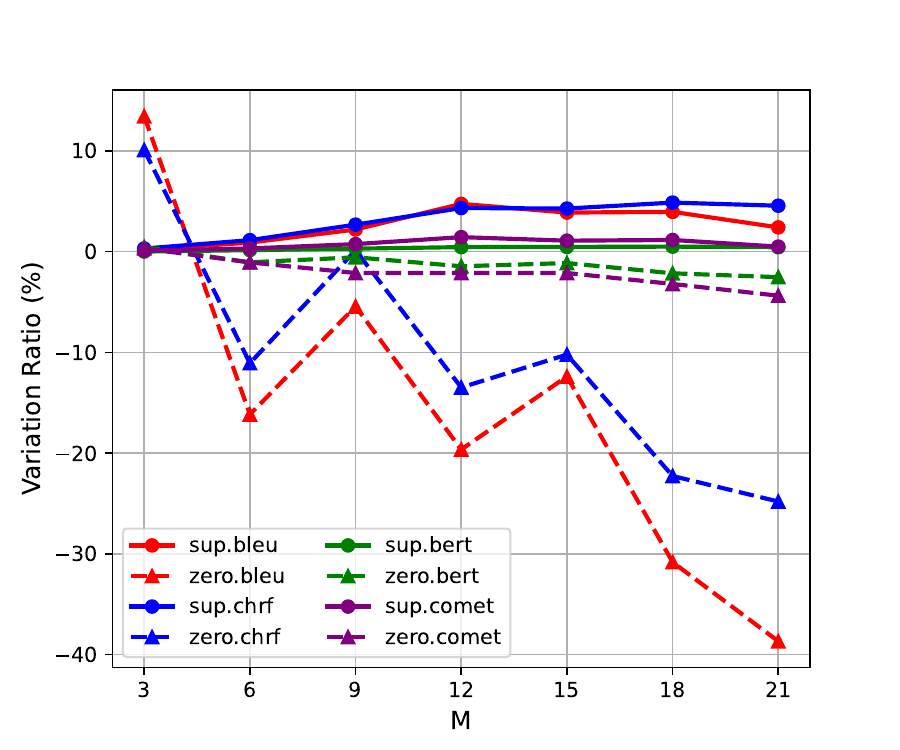}
        \caption{OPUS-100}
        \label{fig:split_opus}
      \end{subfigure}
    \vspace{-1.8em}
    \caption{Variation in different values of M. The y-axis is the variation ratio compared to the performance of the model with prefix decoder-only architecture, and the x-axis is the value of M. The values of $N$ are 6 and 12 in TED-19 and OPUS-100 respectively. Additionally, the line and the dotted line indicate supervised and zero-shot translations respectively. }
    \vspace{-1em}
    \label{fig:split}
\end{figure}
In Section \ref{section:experiments}, we always set $M$ equals $N$ to ensure a fair comparison between the TDO and the encoder-decoder architectures.
However, the balanced design is not optimal \cite{deepencoder, lsls-2023}.
Thus, we test different $M$ on TED-19 and OPUS-100 to investigate balancing two stages.
As shown in Figure \ref{fig:split_ted}, the performance is always improved with the increase of $M$ on TED-19.
On OPUS-100, as depicted in Figure \ref{fig:split_opus}, the case with $M=3$ achieves the best zero-shot translation scores, but there is a noticeable decline in zero-shot translation performance with the increase of $M$, although supervised translation scores continue to rise.

Those results align with our expectations.
As shown in Table \ref{tab:result1}: 1) models with the decoder-only architecture consistently underperform compared to those with the encoder-decoder architecture in supervised translation; 2) models with the decoder-only architecture underperform in zero-shot translation on TED-19 but outperform on OPUS-100.
Moreover, based on the trends in Figure \ref{fig:split_opus}, we can state that the first stage enhances language transfer but at the cost of learning linguistic diversity, and the second stage benefits linguistic diversity.
This statement aligns with \citet{decoderonly_2022} and is further proven by Table \ref{tab:result1} where incorporating InstruCL can significantly improve the performance of zero-shot translation on OPUS-100.
Thus, we conclude that the first stage is crucial in small-scale datasets, whereas the second stage becomes more significant in large-scale datasets.

\subsection{How to set layer index for InstruCL?}\label{section:setindex}
\begin{figure}[t]
    \centering
      \begin{subfigure}[b]{\linewidth}
        \centering
        \includegraphics[width=\linewidth]{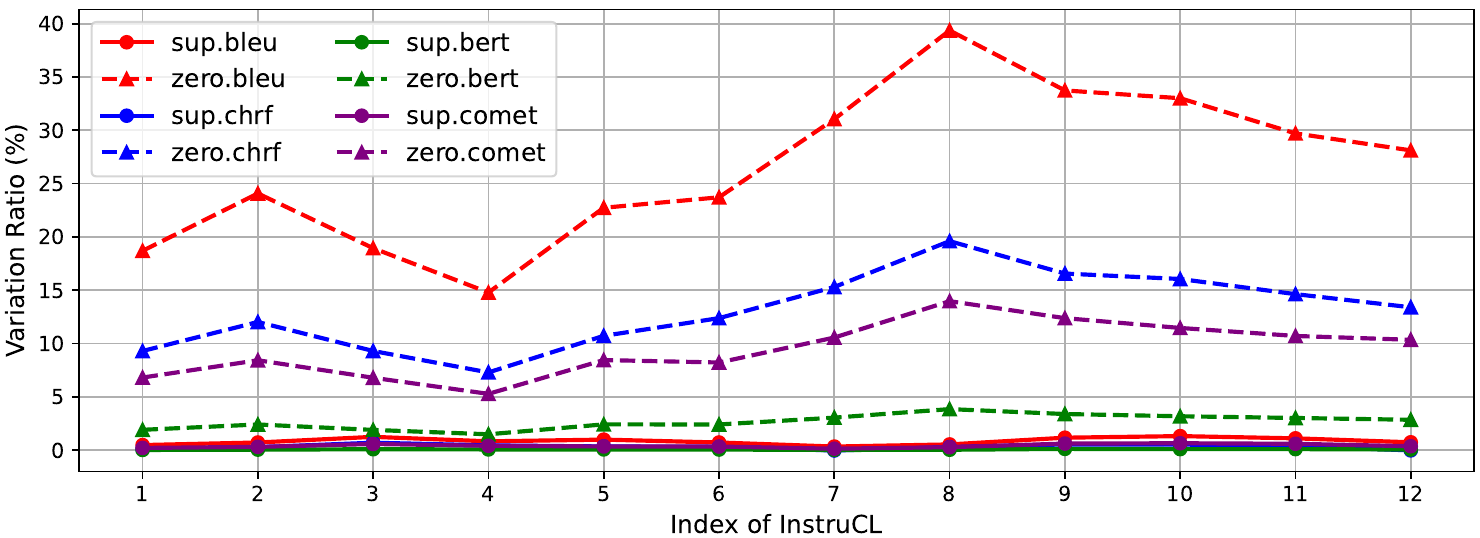}
        \caption{Decoder-only}
        \label{fig:cl_dec}
      \end{subfigure}
      \begin{subfigure}[b]{\linewidth}
        \centering
        \includegraphics[width=\linewidth]{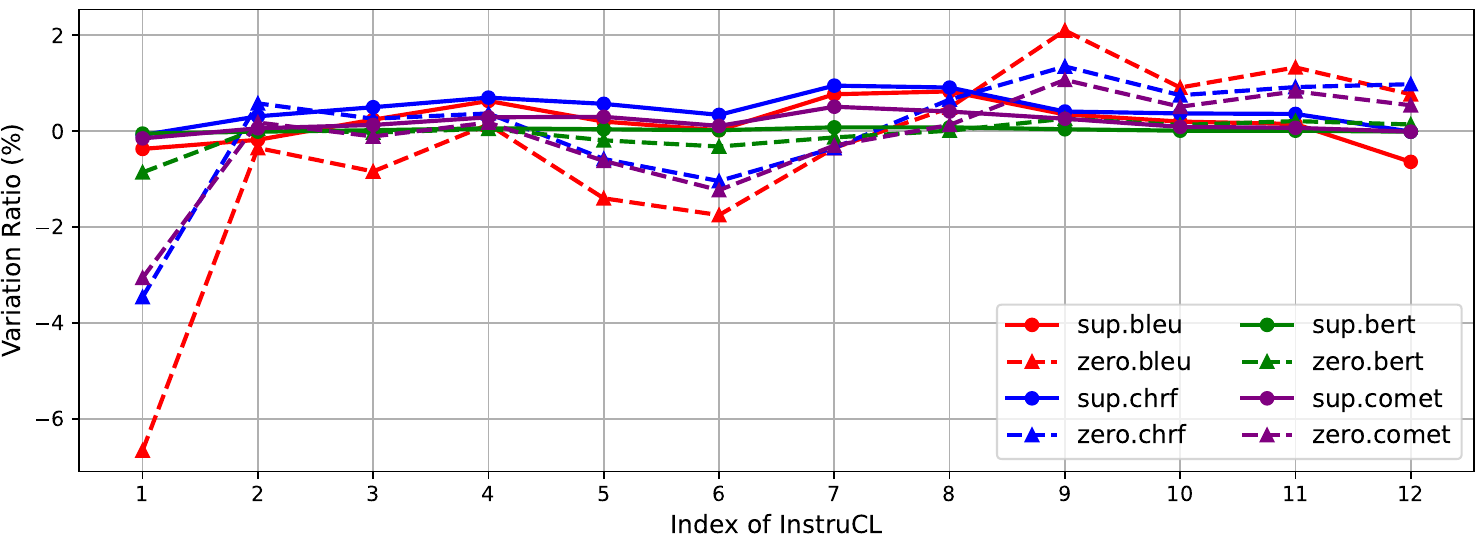}
        \caption{Two-stage Decoder-only with $M=6$}
        \label{fig:cl_sdo}
      \end{subfigure}
    \vspace{-1.8em}
    \caption{Variation in different layer index of InstruCL.
    The y-axis is the variation ratio compared to the performance of the model without InstruCL, and the x-axis is the index of the layer where InstruCL is employed. }
    \vspace{-1em}
    \label{fig:cl}
\end{figure}
In Section \ref{section:experiments}, we set the layer index for InstruCL to 1.5$N$ to prevent the degradation of language transfer in the second stage.
Given that Section~\ref{section:balance} shows the different roles of the first and second stages, we test the performance of models with different layer indexes of InstruCL for the decoder-only and the TDO models.
Figure \ref{fig:cl_dec} demonstrates that InstruCL consistently yields positive gains for the decoder-only architecture.
On the other hand, Figure \ref{fig:cl_sdo} shows a decline in the first stage but benefits in the second stage.
These results indicate that InstruCL primarily affects layers that follow the decoder-only manner, namely, the second stage of TDO, which is further supported by Appendix \ref{appendix:effect}\footnote{Appendix \ref{appendix:effect} shows experiments on implementing InstruCL in different architectures and datasets as a supplement.}.
Moreover, another observation aligning our motivation is that an excessively high index leads to reduced gains.
Therefore, we can conclude that the optimal position for implementing InstruCL is the middle layer of the second stage.

\section{Conclusions}
In this work, we analyzed the reasons behind the underperformance of the decoder-only architecture in MNMT, identifying the lack of language transfer capability as the primary challenge.
To address this, we introduced the Two-stage Decoder-only architecture.
We also proposed Instruction-level Contrastive Learning to overcome the issue from the perspective of representation optimization.
We conducted experiments on two settings, i.e., training from scratch and fine-tuning, using the TED-19 and OPUS-100 datasets, and the results validate the effectiveness of our approach.
Through further experiments and representation analysis, we confirm that the improvements in our methods are derived from enhanced language transfer capabilities.

\section{Limitations}
As mentioned in Section \ref{section:intro}, this work primarily focused on addressing the challenges faced by models with a decoder-only architecture in multilingual neural machine translation (MNMT), rather than exploring how to apply large language models (LLMs), which also have the decoder-only architecture.
This focus is because small models in MNMT still offer the advantages of low training and deployment costs while remaining competitive with LLMs \cite{mnmtVSllm}.
With the increasing interest in improving multilingual translation with LLMs \cite{ftllm}, further exploration is needed to determine whether the representation-level methods proposed in this work can be extended to LLMs.
However, this is beyond the scope of the current study, as the data used to train MNMT models significantly differs from that used to train LLMs.
Therefore, we leave this question for future research.

\section{Ethical Considerations}
All datasets and toolkits used in this work are public, common, and general in the research on multilingual neural machine translation, meanwhile, the usage of those datasets and toolkits follows the license. 
Moreover, this work is foundational research and is not a report of specific applications.
Therefore, this work is harmless and has no ethical risks.

\bibliography{custom}

\appendix

\clearpage
\appendix
\section{Introduction of Illustrating Linguistic Preference}
\label{appendix:preference}
\paragraph{Overview}
In this work, we only quantify the language features of the sentence representation by the similarity scores, although the analysis of \citet{transfer} further quantified the semantic features of representations.
Specifically, the score presents whether the sentence representations at a certain state exhibit more features related to the target language or more features related to the source language.

\paragraph{Setup}
First, quantifying the language features of the sentence representation requires a semantically parallel dataset. 
Therefore, we conduct analysis experiments on TED-19, which provides six fully parallel languages, including \texttt{ar}, \texttt{he}, \texttt{zh}, \texttt{hr}, \texttt{vi}, and \texttt{ja}.
We connect these languages to generate 30 zero-shot translation pairs, each pair consisting of 967 sentences.
The model setup is consistent with our main experiments (Section \ref{section:experiments}).

\paragraph{Computing the similarity score}
First, we follow the process of \citet{transfer} to measure representation similarity in MNMT, employing singular value canonical correlation analysis \cite{svcca}.
As the definition in Section \ref{section:background}, we obtain the token-wise hidden representations of the source sentence, i.e. $\ma{H}$, from a translation pair.
Notably, for a decoder-only model, we cut out the source part, namely, $\vert \ma{H} \vert$ is always $I+1$.
Then, we derive the sentence-level representation $\overline{\ve{h}}$ using average pooling 
$\overline{\ve{h}} = \frac{\sum_{i=1}^q \ve{h}_i}{q}$.
Given $\ma{H}^a$ and $\ma{H}^b$ derived from two sentences, we first perform singular value decomposition on $\overline{\ve{h}}^a$ and $\overline{\ve{h}}^b$ to obtain subspace representations ${\overline{\ve{h}}^{a}} \in \mathbb{R}^{d^a}$ and $ {\overline{\ve{h}}^{b}}  \in \mathbb{R}^{d^b}$.
Then we perform canonical correlation analysis to determine $\mathbf{W}^a \in \mathbb{R}^{d' \times d^a}$ and $\mathbf{W}^b \in  \mathbb{R}^{d' \times d^b}$.
Formally, we compute correlation $\rho$ between $\overline{\ve{h}}^a$ and $\overline{\ve{h}}^b$ as
\begin{equation}
\label{eq:svcca}
% <> inner product
\rho =
\frac{
\langle 
\mathbf{W}^a {\overline{\ve{h}}^a},
\mathbf{W}^b {\overline{\ve{h}}^b} \rangle}
{\Vert 
\mathbf{W}^a {\overline{\ve{h}}^a} 
\Vert \Vert
\mathbf{W}^b {\overline{\ve{h}}^b} 
\Vert},
\end{equation}
where $\langle \cdot, \cdot \rangle$ indicates the inner product. We use $\rho$ to represent the similarity of two sentences.
Subsequently, we get the similarity $\rho_{\ve{x}}$ between $(l_{\ve{y}}, \ve{x}, \ve{y})$ and $(l_{\ve{x}}, \ve{x}, \ve{x})$ and the similarity $\rho_{\ve{y}}$ between $(l_{\ve{y}}, \ve{x}, \ve{y})$ and $(l_{\ve{y}}, \ve{y}, \ve{y})$, respectively.
Therefore, a similarity score of linguistic preference is computed as follows:
\begin{equation}
\label{eq:scores}
s_{(l_{\ve{y}}, \ve{x}, \ve{y})} = \frac{\rho_{\ve{y}}}{\rho_{\ve{y}} + \rho_{\ve{x}}},
\end{equation}
where $s_{(l_{\ve{y}}, \ve{x}, \ve{y})}$ is the similarity score for the given translation pair.
Finally, we compute the set-level score by taking the average scores of all sentences over the test set.

\paragraph{Meaning of the similarity score}
Equation \ref{eq:scores} simply compares the importance of source information and target information in the representation.
Therefore, a value higher than 0.5 means the representation prefers the target language, otherwise the representation prefers the source language.
Moreover, the value reflects the degree of linguistic preference, for example, compared to 0.6, 0.7 means the representation presents much more features of the target language or fewer features of the source language.
In addition, we also denote the highest and lowest values by the vertical lines on each point in Figures \ref{fig:preference} and \ref{fig:representation} to show the value range, which can present stability.
Finally, we can find that models with decoder-only architecture cannot align the representation of the source tokens in the representational subspace of the target language, and they try to align source and target languages to be a language-agnostic state.

\section{Comparison between Different Instruction Strategies in MNMT}
\label{appendix:strategy}
\begin{figure}[t]
    \centering
      \begin{subfigure}[b]{0.49\linewidth}
        \centering
        \includegraphics[width=\linewidth]{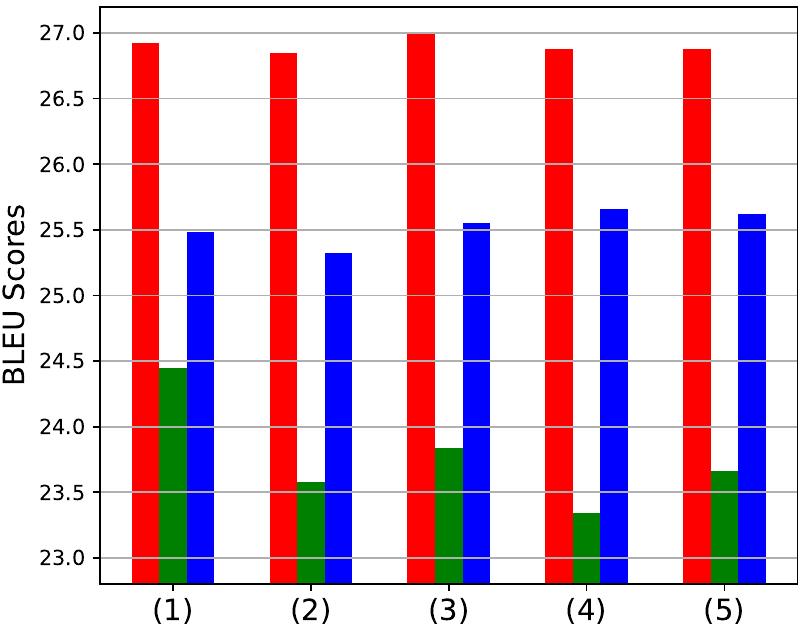}
        \caption{supervised}
        \label{fig:strategy_supervised}
      \end{subfigure}
      \begin{subfigure}[b]{0.49\linewidth}
        \centering
        \includegraphics[width=0.98\linewidth]{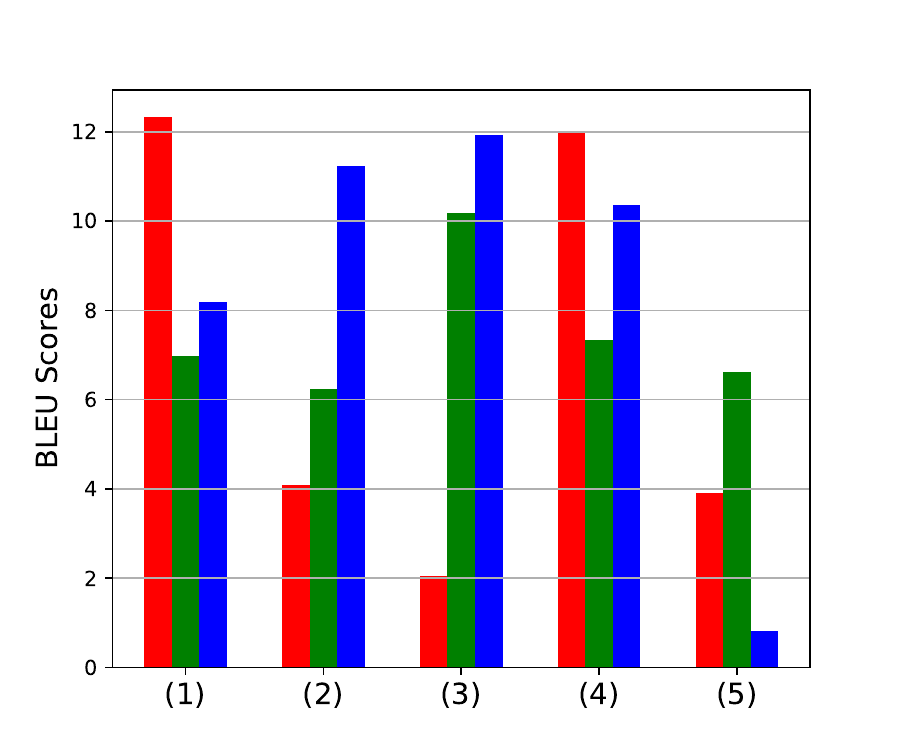}
        \caption{zero-shot}
        \label{fig:strategy_zero}
      \end{subfigure}
    \caption{Averaged BLEU scores in different architectures. The palette follows Figure \ref{fig:comparison}, i.e., red is encoder-decoder, green is causal decoder-only, and blue is prefix decoder-only. }
    \label{fig:strategy}
\end{figure}
MNMT is sensitive to the strategy of translation instruction \cite{TagMatter_2021}.
We summarize the possible strategies as follows: (1) Adding a language tag specified to the target language at the beginning of source tokens; (2) Adding a language tag specified to the target language at the beginning of target tokens; (3) Based on the (2), using the language tag to replace the [eos] token, which is used to be the trigger of inference; (4) Adding two language tag specified to the target language at the beginning of source tokens and the beginning of target tokens, simultaneously; (5) Adding a language tag specified to the source language and a language tag specified to the target language at the beginning of source tokens and target tokens, respectively.
Then, we conduct preliminary experiments on three architectures: encoder-decoder, causal decoder-only, and prefix decoder-only, to support the validity of using approach (1).
As shown in Figure \ref{fig:strategy}, the performance of encoder-decoder architecture meets the analysis of \citet{TagMatter_2021}.
However, a language tag at the beginning of target tokens, i.e., (2), (3), and (4), is more beneficial for the zero-shot capability in Decoder-only architecture.
Considering that (1) also benefits decoder-only architectures in the supervised translation, using (1) in this work is reasonable.

\section{Different Input and Output Forms}
\label{appendix:input}
\begin{figure}[t]
    \centering
    \includegraphics[width=\linewidth]{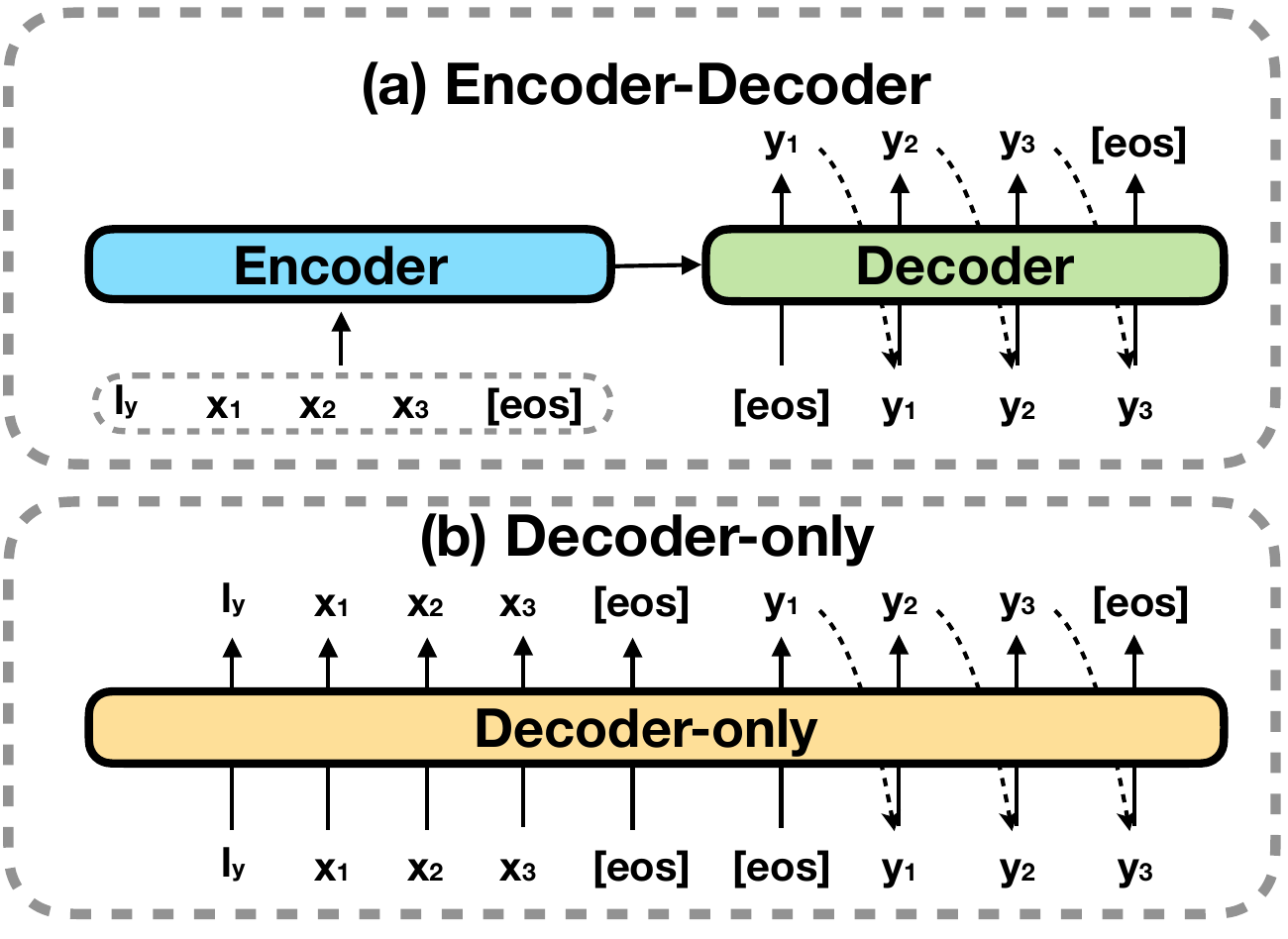}
    \caption{Illustration of input and output forms in MNMT. Subfigures are for the encoder-decoder architecture and the decoder-only architecture, respectively. [eos] is a special token, which means the end of a sentence and is regarded as a token of $\ve{x}$ and $\ve{y}$.}
    \label{fig:input}
    \vspace{-1em}
\end{figure}
Figure \ref{fig:input} illustrates input and output forms for two architectures involved in this work.
Initially, within the encoder-decoder architecture, the encoder receives parallel input from source tokens, including $l_{\ve{y}}$, $\ve{x}$, and a special token [eos].
As a supplement of Section \ref{section:architectures}, for the $I+1$ tokens feeding to the encoder, $l_{\ve{y}}$ is the first token and corresponds to the $\ve{h}_{1}$, then, each index of $\ve{x}$ is shifted, namely, $\ve{x}$ corresponds to $\seq{\ve{h}_{2},...,\ve{h}_{I+1}}$.
Furthermore, the input of the decoder is shifted.
Specifically, in training, [eos] is placed at the beginning of the target tokens, and the output at each position always points to the token in the next position; in inference, [eos] serves as the trigger, and the model would generate the next token step by step until the predicted token is [eos].
Finally, the output of the encoder-decoder architecture only includes target tokens, i.e., $\ve{y}$.
On the other hand, the decoder-only architecture combines source tokens and target tokens as the input.
In this work, we follow \citet{decoderonly_2022, decoderonlymt} to employ MNMT loss instead of language modeling loss, namely, cutting off the source tokens and saving the target tokens only in the ouput,

\section{Attention Mechanisms of Decoder-Only Architectures}
\label{appendix:attention}
\begin{figure}[t]
    \centering
    \includegraphics[width=0.6\linewidth]{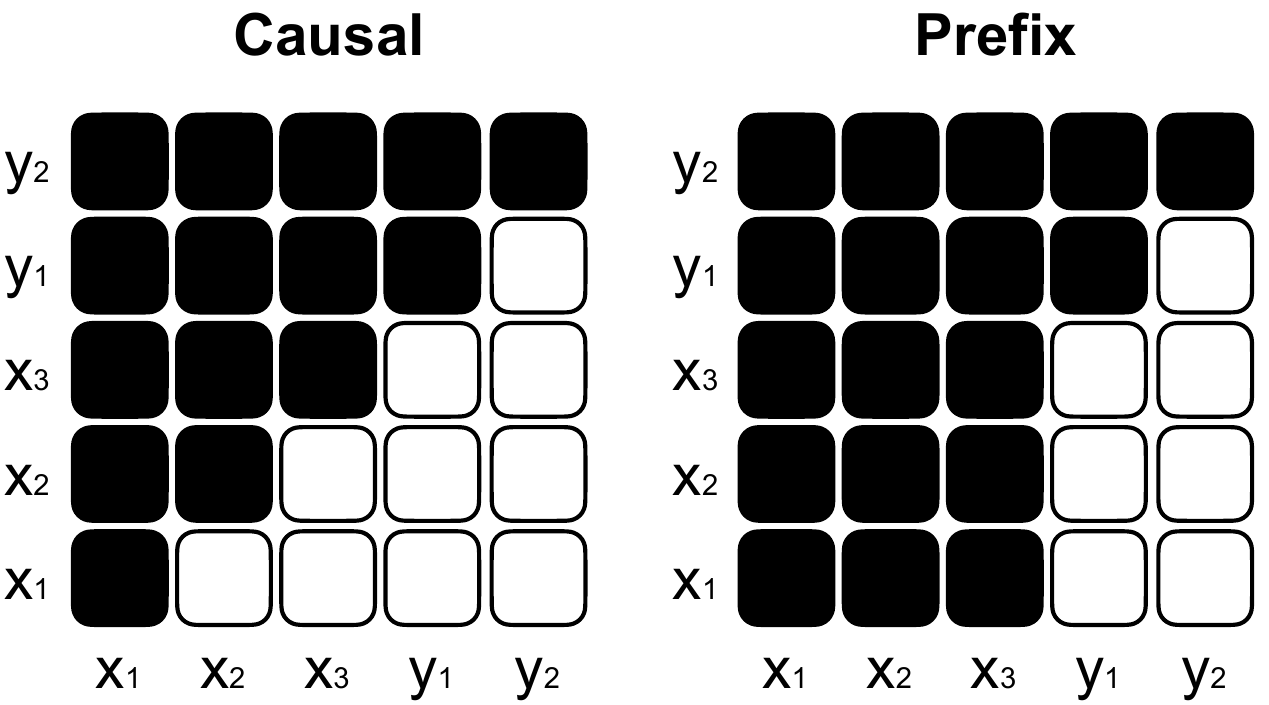}
    \caption{Different manners of the masked self-attention mechanism in the decoder-only architectures. Black blocks mean visible and white blocks mean masked. Thus, source tokens are masked in the causal decoder-only while are visible in the prefix decoder-only. }
    \label{fig:attention}
\end{figure}

As illustrated in Figure \ref{fig:attention}, the causal attention mechanism in the decoder-only architecture treats source and target tokens equally, meaning that each token is influenced solely by preceding tokens and itself.
In contrast, the prefix attention mechanism maintains bi-directional attention for source tokens where source tokens are influenced by each other, while target tokens use mono-directional attention, meaning they are influenced only by prior tokens and themselves.

\section{Estimation of Parameters}
\label{appendix:estimation}
We follow the notation in Section \ref{section:setup}, that is, $d$ is the dimension of the model and the inner size of FFN is 4$d$.
Therefore, each attention mechanism has $4d^2$ parameters because there are 4 matrices with dimensions of $d\times d$, and each FFN has $8d^2$ parameters \cite{transformer}.
Then, all layers have the structure illustrated in Figure \ref{fig:background}. 
Given $N=$1, the model with encoder-decoder architecture has $28d^2$ parameters and the model with Decoder-only architecture has $24d^2$ parameters.
Thus, considering the fixed parameters of normalization modules and embedding layer, Decoder-only architecture is implemented with around 10\% fewer parameters than encoder-decoder architecture.

\section{Detailed Information of Datasets}
\label{appendix:dataset}
First, the language codes used in our descriptions adhere to ISO 639-1\footnote{\url{https://www.loc.gov/standards/iso639-2/php/code_list.php}}. As described in Section \ref{section:dataset}, the first dataset is TED-19 \cite{transfer}, a subset of TED Talks \cite{ted} containing 6.5 million instances across 19 languages from various language families. This dataset includes 32 supervised translation pairs and 306 zero-shot translation pairs. Detailed information about TED-19 is provided in Table \ref{tab:ted}.
The second dataset is the revised version of OPUS-100 \cite{massive-2020, TLP-2021}, which includes 95 languages and a total of 92 million instances. Notably, the zero-shot translation in OPUS-100 involves only six languages (\texttt{ar}, \texttt{nl}, \texttt{de}, \texttt{zh}, \texttt{ru}, and \texttt{fr}), resulting in 30 translation pairs. Additionally, we further cleaned the dataset by removing noisy instances containing unreadable characters, even though \citet{TLP-2021} had already removed repetitions from the original OPUS-100 dataset \cite{massive-2020}. Detailed information about OPUS-100 can be found in Table \ref{tab:opus}.
Generally, each pair of validation and test sets in these two datasets contains 2,000 instances, though several pairs in OPUS-100 have fewer instances. Finally, we used SentencePiece \cite{sentencepiece} to generate the vocabulary for training, with the vocabulary size set to 50,000 for TED-19 and 64,000 for OPUS-100.

\begin{table*}[!ht]
    \centering
    \resizebox{\textwidth}{!}{
    \begin{tabular}{l cccccccccccccccc}
    \toprule
        ~ & ~ & ~ & ~ & ~ & \multicolumn{3}{c}{BLEU $\uparrow$ } & \multicolumn{3}{c}{chrF++ $\uparrow$ } & \multicolumn{3}{c}{BERTScore $\uparrow$ }&\multicolumn{3}{c}{COMET $\uparrow$ } \\ 
    \midrule
        ~ & ~ & \#enc & \#dec & idx. & \texttt{en}$\to$ & $\to$\texttt{en} & zero & \texttt{en}$\to$ & $\to$\texttt{en} & zero & \texttt{en}$\to$ & $\to$\texttt{en} & zero & \texttt{en}$\to$ & $\to$\texttt{en} & zero\\ 
    \midrule
        \multirow{8}{*}{\begin{minipage}{0.9cm} TED\\ $d=$512 \end{minipage}} & \multirow{2}{*}{Enc-Dec} & 6 & 6 & - & 25.46  & 28.31  & 12.32  & 45.96  & 50.86  & 32.13  & 84.10  & 93.37  & 78.03  & 80.49  & 78.15  & 67.26 \\
        ~ & ~ & 6 & 6 & 6 &  24.92 & 28.39 & 12.96 & 45.56 & 50.97 & 33.42 & 83.94 & 93.68 & 79.10 & 79.99 & 78.21 & 70.37 \\ 
        \cdashline{3-17}
        ~ & \multirow{3}{*}{Dec-only} & 0 & 12 & - & 24.00  & 26.97  & 8.18  & 44.49  & 48.93  & 25.35  & 83.54  & 92.97  & 74.52  & 78.46  & 76.10  & 56.74\\  
        ~ & ~ & 0 & 12 & 6 & 24.16 & 27.18 & 10.12 & 44.61 & 49.11 & 28.49 & 83.63 & 93.01 & 76.32 & 78.80 & 76.30 & 61.41 \\
        ~ & ~ & 0 & 12 & 9 & 24.26 & 27.31 & 10.94 & 44.69 & 49.24 & 29.55 & 83.69 & 93.05 & 77.05 & 79.08 & 76.45 & 63.77 \\
        \cdashline{3-17}
        ~ & \multirow{3}{*}{TDO} & 0 & 12 & - & 25.53  & 28.76  & 14.26  & 46.01  & 51.09  & 34.72  & 84.13  & 93.41  & 79.27  & 80.43  & 78.18  & 70.82 \\ 
        ~ & ~ & 0 & 12 & 6 &  25.46 & \textbf{29.02} & 14.06 & 45.98 & 51.44 & 34.38 & 84.10 & 93.48 & 79.15 & 80.47 & 78.54 & 70.51\\ 
        ~ & ~ & 0 & 12 & 9 & \textbf{25.62}  & 28.94  & \textbf{14.70}  & \textbf{46.15}  & \textbf{51.46}  & \textbf{35.34}  & \textbf{84.15}  & \textbf{93.47}  & \textbf{79.57}  & \textbf{80.55}  & \textbf{78.55}  & \textbf{71.94}  \\ 
    \midrule
        \multirow{8}{*}{\begin{minipage}{0.9cm} OPUS\\ $d=$512 \end{minipage}} & \multirow{2}{*}{Enc-Dec} & 12 & 12 & - & \textbf{25.18}  & 29.79  & 5.13  & \textbf{44.75}  & 48.40  & 12.95  & \textbf{82.98}  & 92.33  & 72.44  & \textbf{76.59}  & 76.21  & 58.51\\ 
         ~ & ~ & 12 & 12 & 12 & 24.98 & 29.61 & 6.56 & 44.65 & 48.30 & 15.49 & 82.97 & 92.34 & 73.45 & 76.46 & 76.23 & 59.61\\ 
        \cdashline{3-17}
        ~ & \multirow{3}{*}{Dec-only} & 0 & 24 & - & 23.96  & 28.41  & 6.62  & 42.98  & 47.22  & 15.36  & 82.47  & 92.06  & 73.57  & 75.48  & 75.34  & \textbf{59.56} \\ 
        ~ & ~ & 0 & 24 & 12&24.22 & 28.26 & 6.99 & 43.23 & 46.83 & 15.98 & 82.49 & 92.04 & 73.66 & 75.55 & 74.94 & 59.42 \\
        ~ & ~ & 0 & 24 & 18 &23.98 & 28.22 & 6.73 & 43.18 & 46.80 & 16.17 & 82.52 & 92.07 & 73.67 & 75.60 & 75.12 & 59.37\\
        \cdashline{3-17}
        ~ & \multirow{3}{*}{TDO} & 0 & 24 & - & 24.88  & \textbf{29.97}  & 5.32  & 44.72  & \textbf{49.39}  & 13.29  & 82.91  & \textbf{92.41}  & 72.50  & 76.26  & \textbf{76.73}  & 58.30 \\ 
        ~ & ~ & 0 & 24 & 12 &24.61 & 29.37 & 6.46 & 44.68 & 48.72 & 15.14 & 82.87 & 92.37 & 73.30 & 76.16 & 76.21 & 59.41\\
        ~ & ~ & 0 & 24 & 18 & 24.35  & 29.52  & \textbf{7.93}  & 44.44  & 48.74  & \textbf{18.65}  & 82.84  & 92.37  & \textbf{73.97}  & 75.93  & 76.23  & 58.71 \\ 
    \midrule
        \multirow{8}{*}{\begin{minipage}{0.9cm} OPUS\\ $d=$1024 \end{minipage}} & \multirow{2}{*}{Enc-Dec} & 6 & 6 & - & 27.71  & \textbf{31.60}  & 6.95  & 46.84  & 50.31  & 15.89  & 83.55  & 92.62  & 74.12  & \textbf{78.10}  & 77.58  & 59.99\\ 
        ~ & ~ & 6 & 6 & 6 & \textbf{27.74} & 31.52 & 7.75 & \textbf{46.92} & 49.91 & 18.06 & \textbf{83.56} & \textbf{92.66} & 74.44 & 78.07 & 77.69 & 60.43\\ 
        \cdashline{3-17} 
        ~ & \multirow{3}{*}{Dec-only} & 0 & 12 & - & 26.79  & 30.42  & 8.15  & 45.48  & 48.92  & 17.65  & 83.21  & 92.37  & 74.17  & 77.53  & 76.69  & \textbf{62.32}  \\ 
        ~ & ~ & 0 & 12 & 6 & 26.87 & 30.72 & 8.47 & 45.58 & 49.18 & 17.78 & 83.53 & 92.51 & 74.38 & 77.74 & 77.82 & 61.61 \\
        ~ & ~ & 0 & 12 & 9 & 26.72 & 30.09 & 8.42 & 45.34 & 48.52 & 17.33 & 83.16 & 91.83 & 74.23 & 77.31 & 76.61 & 61.55 \\
        \cdashline{3-17}
        ~ & \multirow{3}{*}{TDO} & 0 & 12 & - & 27.22  & 31.58  & 7.06  & 46.54  & \textbf{50.59}  & 15.96  & 83.44  & 92.64  & 73.78  & 77.68  & \textbf{77.89}  & 60.60 \\ 
        ~ & ~ & 0 & 12 & 6 & 26.72 & 31.05 & 7.43 & 45.49 & 49.54 & 16.25 & 83.19 & 92.40 & 74.00 & 77.45 & 77.49 & 61.89 \\
        ~ & ~ & 0 & 12 & 9 & 27.12  & 31.49  & \textbf{9.28}  & 46.55  & 50.23  & \textbf{21.33}  & 83.50  & 92.65  & \textbf{75.04}  & 77.63  & 77.64  & 60.84  \\ 
    \bottomrule
    \end{tabular}}
    \caption{Averaged scores of results in experiments of training from scratch and verifying InstruCL across different architectures.
    Both the decoder-only and TDO architectures adopt the prefix attention mechanism.
    All terms, settings, and abbreviations follow the Table \ref{tab:result1}.
    Moreover, \#enc, \#dec, and idx. indicate the number of encoder layers, the number of decoder layers, and the layer index where to implement InstruCL, respectively.
    In addition, the placeholder (-) in the collum of idx. means that InstruCL is not implemented in this row.
    The best score in each column and block is in bold.}
  \label{tab:result_cl}
\end{table*}

\section{Selection Standards of Evaluation Metrics}\label{appendix:standards}
First, SacreBLEU \cite{sacrebleu}, an implementation of BLEU \cite{bleu}, measures the lexical overlap between generated translations and reference translations.
chrF++ evaluates overlap at the character level and accounts for a balance between precision and recall. These two metrics can corroborate each other's results.
On the other hand, BERTScore\footnote{For BERTScore, \texttt{en} is computed using \textit{xlmr.large} \cite{roberta, xlmr}, while other languages are computed using \textit{bert-base-multilingual-cased} \cite{bert}.} \cite{bertscore} measures the similarity between generated translations and references at the representation level.
COMET\footnote{All COMET scores are computed using \textit{Unbabel/wmt22-comet-da} \cite{comet-22}.} \cite{comet} also evaluates representational similarity, with an additional emphasis on the source text for enhanced semantic relevance.
Intuitively, BERTScore may penalize instances that do not translate into the expected target language, while COMET is more sensitive to semantic relevance.
To validate this intuition, we also introduce the target-off ratio as a secondary evaluation metric.
Notably, it is considered secondary because the testing tools are not entirely accurate, particularly when recognizing low-resource languages, as they rely on language-specific tokens.

\section{Detailed Model Settings}\label{appendix:model}
We implement models by Fairseq \cite{fairseq}, an open-source toolkit.
First of all, in this work, we apply independent sinusoidal positional embeddings for source tokens and target tokens \cite{transformer} for the input of the decoder-only architecture.
Notably, the estimation of parameters in modeling is introduced in Appendix \ref{appendix:estimation}.

\paragraph{Model settings of training from scratch}
In the case of training from scratch on TED-19, we set $N$ to 6, $d$ to 512, inner size of FFN to 4$d$.
Thus, the model with an encoder-decoder architecture has 70 million parameters, while the model with a decoder-only architecture has 63 million parameters.
Moreover, the FFN in the adaptation module matches the dimensions of the FFN in the main part, so in this case, the model has 67 million parameters.
In the training, we set the learning rate to 0.0005 and the model is trained for 30 epochs on eight NVIDIA V100 GPUs with a batch size of 4,000 per GPU to ensure full convergence.
Moreover, we set the head number of the attention mechanism to 8, the dropout rate to 0.1, label smoothing to 0.1, and weight decay to 0.0001.
We also employ Adam \cite{adam} as our optimizer and set \textit{share-all-embeddings} of Fairseq.
We evaluate by averaging the top-5 best checkpoints selected based on validation loss.
In the case of training from scratch on OPUS-100, we first increase $N$ to 12, resulting in parameter counts of 121 million, 108 million, and 113 million, respectively.
In the training, we set the number of gradient accumulation steps to 16 to increase the batch size and train for 50,000 steps with a learning rate of 0.0007.
We also consider a wider model where $N$ is 6, $d$ is 1024, and the head number of the attention mechanism is 16, resulting in parameter counts of 242 million, 217 million, and 234 million, respectively.
When, we additionally set an attention dropout to 0.05 and reduce the learning rate to 0.0005 for a stable gradient.
Moreover, we reduce the batch size per GPU to 2,000, set the number of gradient accumulation steps to 32, and train for 100,000 steps due to GPU memory constraints.
For two cases of OPUS-100, we test the checkpoint with the best validation loss.
Additionally, in training on OPUS-100, we set \textit{encoder-normalize-before} and \textit{decoder-normalize-before} in Fairseq and reduce the weight decay to 0, which lead to a quick convergence in a complex data condition \cite{mbart,m2m, nllb}.

\paragraph{Model settings of fine-tuning}
In the model settings of fine-tuning, M2M-418M has 12 layers for encoder and decoder, respectively, where $d$ of M2M-418M is 1024, and the inner size of FFN is 4096, based on the description in Section \ref{section:setup}, we set $N$ to 6, resulting in parameter counts of 307 million, 282 million, and 299 million, respectively.
In the training, the label smoothing is 0.2, the dropout is 0.3, the attention dropout is 0.05, and the batch size and the learning rate keep the settings of training from scratch.
Then, given that NLLB-600M has the same configuration as M2M-418M but with a larger vocabulary size, the same setting of hyper-parameters leads to the count of parameters increased to 439 million, 413 million, and 430 million, respectively, and, we reduce the batch size to 2000 and set gradient accumulation to 2 for NLLB-600M because of the GPU memory constraints.
In M2M-1.2B, which has 24 decoder layers and a larger inner size of FFN compared to M2M-418M, we set $N$ to 12, leading to parameter counts of 685 million, 635 million, and 668 million, respectively, and our experiments are conducted on four NVIDIA A6000 GPUs, and we set gradient accumulation to 2.
We also reduce the learning rate to 0.0002 and the number of training epochs to 10 because of more parameters.

\section{The Effectiveness of InstruCL on Encoder-Decoder Architecture}
\label{appendix:effect}
As a supplementary trail for Sections \ref{section:result1} and \ref{section:setindex}, we conduct experiments on applying InstruCL to the encoder-decoder, the prefix decoder-only, and TDO architectures, and then compare their performances on three cases of training from scratch described in Section \ref{section:setup}.
The layer index where InstruCL is implemented at the TDO is 1.5$N$.
We also implement InstruCL for the decoder-only architecture at the same layer as a comparison.
However, given that the number of encoder layers in an encoder-decoder architecture is $N$, InstruCl is implemented at the output of the encoder, namely, the layer index is $N$.
Therefore, as comparison groups, we also implement InstruCL for the decoder-only and TDO architectures at the $N$ layer.

Tabel \ref{tab:result_cl} shows the experimental results.
The first observation is that the encoder-decoder architecture can be gained from InstruCL due to the improved performance in all cases.
Notably, the first observation is not violated from the statement in Section \ref{section:setindex} that InstruCL mainly affects the layer following the decoder-only manner, because of the performance of TDO in TED-19 and OPUS-100.
Specifically, considering the decoder-only architecture, first, in the TED-19, when the index is set to $N$, Dec-only shows a significant improvement in zero-shot translations with BLEU scores increasing by 1.94, while TDO degraded by 0.64.
Second, in two cases from the OPUS-100, when the index is set to $1.5N$, TDO achieves significant improvements of 2.61 and 2.22, respectively.
Third, in three cases, compared to setting the index to $N$, the decoder-only model showed smaller gains or even degradations when the index is set to $1.5N$, with scores increasing by 0.82, -0.26, and -0.05.

These results are consistent with our statement in Section \ref{section:instrucl}.
Specifically, the first stage of TDO overlaps with InstruCL in terms of facilitating the learning of target language representations, which explains the suboptimal performance when both are used together.
Additionally, InstruCL is most effective when applied in the middle layers, which align with the decoder-only manner.
On the other hand, considering the performance of the vanilla models, i.e., Enc-Dec and Dec-only, we can assert that InstruCL, which does not require additional data costs, generally benefits all architectures.

\begin{table}[t]
    \centering
    \resizebox{\linewidth}{!}{
    \begin{tabular}{lcccccc}
    \toprule
         & $d$ & $d^{1}_{\text{ffn}}$ & $d^{2}_{\text{ffn}}$ & \texttt{en}$\to$ & $\to$\texttt{en} & zero \\ 
    \midrule
        TDO+adapt. & 512 & 2048 & 2048 & \textbf{25.61}  & 28.52  & \textbf{14.51}  \\ 
    \hdashline
        \multirow{4}{*}{TDO} & 544 & 2048 & 2048 & 25.55  & 28.28  & 14.22  \\ 
        ~ & 512 & 2432 & 2432 & 25.51  & 28.51  & 14.31  \\ 
        ~ & 512 & 2048 & 2816 & 25.32  & 27.98  & 13.89  \\ 
        ~ & 512 & 2816 & 2048 & 25.56  & \textbf{28.95}  & 14.01  \\ 
    \bottomrule
    \end{tabular}
    }
    \caption{Averaged BLEU scores of models with TDO architecture trained on TED-19.
    Abbreviations in this table follow Table \ref{tab:result1}.
    In addition, $d^{1}_{\text{ffn}}$ is the inner size of FFN in the first stage, and $d^{2}_{\text{ffn}}$ is in the second stage.
    The best score is in bold.
    }
    \label{tab:params}
\end{table}

\section{Adaption Modules Do Not Equal Simply Increasing Parameters}
\label{appendix:params}
Adding adaptation modules increases the number of parameters, so it is crucial to determine whether the gains from these modules are primarily due to the increased parameters. As shown in Table \ref{tab:params}, we directly increased the parameters of the TDO model using various strategies, ensuring that the number of parameters is comparable to or even greater than that of the TDO model with adaptation modules.
The results demonstrate that the TDO model with adaptation modules outperforms in zero-shot translation and in translating supervised pairs from English to non-central languages. Notably, considering the previous point, the reason why adaptation modules do not achieve the best performance when translating from non-central languages to English can be attributed to their effectiveness in preventing overfitting of English, which dominates the multilingual representations due to most of the training data being in English \cite{gu-2019, adaptingZero}.
Therefore, the results in this table support our assertion that the gains from adaptation modules cannot be simply attributed to increasing parameters.

\begin{table*}[!ht]
    \centering
    \resizebox{\textwidth}{!}{
    \begin{tabular}{cccccccccc}
    \toprule
        Code & Language & Family & Sub-Family & \#Train & Code & Language & Family & Sub-Family & \#Train \\ 
    \midrule
        es & Spanish & Indo-European & Romance & 196026 & ar & Arabic & Afro-Asiatic & Semitic & 214111 \\ 
        fr & French & Indo-European & Romance & 192304 & he & Hebrew & Afro-Asiatic & Semitic & 211819 \\ 
        ro & Romanian & Indo-European & Romance & 180484 & ru & Russian & Indo-European & Slavic & 208458 \\ 
        nl & Dutch & Indo-European & Germanic & 183767 & ko & Korean & Koreanic & ~ & 205640 \\ 
        de & German & Indo-European & Germanic & 167888 & it & Italian & Indo-European & Romance & 204503 \\ 
        pl & Polish & Indo-European & Slavic & 176169 & ja & Japanese & Japonic & ~ & 204090 \\ 
        hr & Croatian & Indo-European & Slavic & 122091 & zh & Chinese & Sino-Tibetan & Sinitic & 199855 \\ 
        cs & Czech & Indo-European & Slavic & 103093 & tr & Turkish & Turkic & ~ & 182470 \\ 
        fa & Persian & Indo-European & Iranian & 150965 & vi & Vietnamese & Austroasiatic & Vietic & 171995 \\ 
    \bottomrule
    \end{tabular}
    }
    \caption{Detailed information of TED-19 datasets. \#Train indicates the number of training instances.}
    \label{tab:ted}
\end{table*}

\begin{table*}[!ht]
    \centering
    \resizebox{\textwidth}{!}{
    \begin{tabular}{cccccccccc}
    \toprule
        Code & Language & Family & Sub-Family & \#Train & Code & Language & Family & Sub-Family & \#Train \\ 
    \midrule
        fa & Persian & Indo-European & Iranian & 934413 & yi & Yiddish & Indo-European & Romance & 1865 \\ 
        bn & Bengali & Indo-European & Iranian & 724719 & ga & Irish & Indo-European & Celtic & 187967 \\ 
        ur & Urdu & Indo-European & Iranian & 724226 & br & Breton & Indo-European & Celtic & 96951 \\ 
        si & Sinhala & Indo-European & Iranian & 613702 & cy & Welsh & Indo-European & Celtic & 92615 \\ 
        hi & Hindi & Indo-European & Iranian & 374472 & gd & Scottish Gaelic & Indo-European & Celtic & 11104 \\ 
        tg & Tajik & Indo-European & Iranian & 183216 & lt & Lithuanian & Indo-European & Baltic & 797693 \\ 
        ne & Nepali & Indo-European & Iranian & 144520 & lv & Latvian & Indo-European & Baltic & 779972 \\ 
        gu & Gujarati & Indo-European & Iranian & 108564 & tr & Turkish & Turkic & ~ & 918838 \\ 
        ku & Kurdish & Indo-European & Iranian & 107110 & az & Azerbaijani & Turkic & ~ & 237533 \\ 
        pa & Punjabi & Indo-European & Iranian & 72160 & uz & Uzbek & Turkic & ~ & 148319 \\ 
        as & Assamese & Indo-European & Iranian & 58009 & tt & Tatar & Turkic & ~ & 97746 \\ 
        mr & Marathi & Indo-European & Iranian & 26117 & ug & Uyghur & Turkic & ~ & 71241 \\ 
        ps & Pashto & Indo-European & Iranian & 14254 & kk & Kazakh & Turkic & ~ & 62227 \\ 
        or & Oriya & Indo-European & Iranian & 13410 & ky & Kyrgyz & Turkic & ~ & 12724 \\ 
        de & German & Indo-European & Germanic & 968252 & tk & Turkmen & Turkic & ~ & 98 \\ 
        nl & Dutch & Indo-European & Germanic & 936611 & ar & Arabic & Afro-Asiatic & Semitic & 959868 \\ 
        sv & Swedish & Indo-European & Germanic & 916259 & he & Hebrew & Afro-Asiatic & Semitic & 913493 \\ 
        no & Norwegian & Indo-European & Germanic & 914187 & mt & Maltese & Afro-Asiatic & Semitic & 672134 \\ 
        da & Danish & Indo-European & Germanic & 911156 & ha & Hausa & Afro-Asiatic & Chadic & 91869 \\ 
        is & Icelandic & Indo-European & Germanic & 813820 & am & Amharic & Afro-Asiatic & Semitic & 64369 \\ 
        nn & Norwegian Nynorsk & Indo-European & Germanic & 172187 & el & Greek & Indo-European & Hellenic & 932811 \\ 
        af & Afrikaans & Indo-European & Germanic & 146600 & sq & Albanian & Indo-European & Albanian & 855095 \\ 
        nb & Norwegian Bokmål & Indo-European & Germanic & 128374 & ml & Malayalam & Dravidian & ~ & 633920 \\ 
        fy & Frisian & Indo-European & Germanic & 42372 & ta & Tamil & Dravidian & ~ & 184699 \\ 
        li & Limburgish & Indo-European & Germanic & 3331 & te & Telugu & Dravidian & ~ & 37792 \\ 
        ru & Russian & Indo-European & Slavic & 951611 & kn & Kannada & Dravidian & ~ & 13777 \\ 
        sr & Serbian & Indo-European & Slavic & 935342 & xh & Xhosa & Niger-Congo & Bantu & 231708 \\ 
        hr & Croatian & Indo-European & Slavic & 927541 & rw & Kinyarwanda & Niger-Congo & Bantu & 62159 \\ 
        pl & Polish & Indo-European & Slavic & 926940 & zu & Zulu & Niger-Congo & Bantu & 6834 \\ 
        bg & Bulgarian & Indo-European & Slavic & 925647 & ig & Igbo & Niger-Congo & Volta-Niger & 691 \\ 
        cs & Czech & Indo-European & Slavic & 924282 & fi & Finnish & Uralic & Finnic & 938601 \\ 
        bs & Bosnian & Indo-European & Slavic & 921232 & et & Estonian & Uralic & Finnic & 893074 \\ 
        sl & Slovenian & Indo-European & Slavic & 912248 & hu & Hungarian & Uralic & Finno-Ugric & 920592 \\ 
        mk & Macedonian & Indo-European & Slavic & 881176 & se & Northern Sami & Uralic & Sami & 32289 \\ 
        sk & Slovak & Indo-European & Slavic & 878540 & vi & Vietnamese & Austroasiatic & Vietic & 883581 \\ 
        uk & Ukrainian & Indo-European & Slavic & 759826 & id & Indonesian & Austronesian & Malayo-Polynesian & 881198 \\ 
        sh & Serbo-Croatian & Indo-European & Slavic & 209379 & ms & Malay & Austronesian & Malayo-Polynesian & 819431 \\ 
        be & Belarusian & Indo-European & Slavic & 61862 & mg & Malagasy & Austronesian & Malayo-Polynesian & 292520 \\ 
        fr & French & Indo-European & Romance & 963140 & km & Khmer & Austroasiatic & Khmeric & 101294 \\ 
        es & Spanish & Indo-European & Romance & 929677 & zh & Chinese & Sino-Tibetan & Sinitic & 954358 \\ 
        it & Italian & Indo-European & Romance & 928427 & my & Burmese & Sino-Tibetan & Lolo-Burmese & 5326 \\ 
        pt & Portuguese & Indo-European & Romance & 919755 & th & Thai & Kra-Dai & Tai & 892433 \\ 
        ro & Romanian & Indo-European & Romance & 913451 & ko & Korean & Koreanic & ~ & 892064 \\ 
        ca & Catalan & Indo-European & Romance & 633826 & ja & Japanese & Japonic & ~ & 886850 \\ 
        gl & Galician & Indo-European & Romance & 353596 & eu & Basque & Language isolate & ~ & 786645 \\ 
        wa & Walloon & Indo-European & Romance & 48894 & eo & Esperanto & Constructed & ~ & 257560 \\ 
        oc & Occitan & Indo-European & Romance & 27773 & ka & Georgian & Kartvelian & ~ & 240335 \\ 
    \bottomrule
    \end{tabular}}
    \caption{Detailed information of OPUS-100 datasets. \#Train indicates the number of training instances.}
    \label{tab:opus}
\end{table*}

\end{document}